%% file: main.tex
\documentclass[letterpaper]{article} % DO NOT CHANGE THIS
\usepackage{aaai24}  % DO NOT CHANGE THIS
\usepackage{times}  % DO NOT CHANGE THIS
\usepackage{helvet}  % DO NOT CHANGE THIS
\usepackage{courier}  % DO NOT CHANGE THIS
\usepackage[hyphens]{url}  % DO NOT CHANGE THIS
\usepackage{graphicx} % DO NOT CHANGE THIS
\usepackage{natbib}  % DO NOT CHANGE THIS AND DO NOT ADD ANY OPTIONS TO IT
\usepackage{caption} % DO NOT CHANGE THIS AND DO NOT ADD ANY OPTIONS TO IT
\DeclareCaptionStyle{ruled}
{labelfont=normalfont,labelsep=colon,strut=off} % DO NOT CHANGE THIS
\usepackage{mdframed}
\newmdenv[
  topline=false,
  bottomline=false,
  rightline=false,
  linewidth=0.3mm,
  skipabove=1mm,
  skipbelow=1mm
]{siderules*}
\usepackage{microtype}
\usepackage{algorithm}
\usepackage{algpseudocode}
\usepackage{subfigure}
\usepackage{booktabs} % for professional tables
\usepackage{centernot}
\usepackage{enumitem}
\usepackage{multicol}
\usepackage[utf8]{inputenc} % allow utf-8 input
\usepackage[T1]{fontenc}    % use 8-bit T1 fonts
\usepackage{booktabs}       % professional-quality tables
\usepackage{amsfonts}       % blackboard math symbols
\usepackage{nicefrac}       % compact symbols for 1/2, etc.    % microtypography
\usepackage{xcolor}         % colors
\usepackage{tikz}
\usepackage{url}
\usepackage{verbatim}
\usepackage{mathtools}
\usepackage{amsthm}
\usepackage{thmtools}
\usepackage{thm-restate}
\usepackage{caption}
\usepackage{enumitem}
\usepackage{booktabs}
\newcommand{\ind}{\perp\!\!\!\!\perp} 
\usepackage{amssymb}
\usepackage{float}
\usepackage{amsmath}
\usepackage{xspace}
\definecolor{r}{rgb}{0, 0, 0}

\newtheorem{proposition-informal}{Proposition (Informal)}[section]
\newtheorem{definition}{Definition}[section]
\newtheorem{assumption}{Assumptions}[section]

\DeclareMathOperator*{\argmin}{arg\,min}
\usepackage{color, colortbl}
\usepackage{mdframed}
\newmdenv[
  topline=false,
  bottomline=false,
  rightline=false,
  linewidth=0.3mm,
  skipabove=1mm,
  skipbelow=1mm
]{siderules}

\definecolor{Gray}{gray}{0.9}
\usepackage{newfloat}
\usepackage{listings}
\floatstyle{ruled}
\newfloat{listing}{tb}{lst}{}
\floatname{listing}{Listing}
\title{NESTER: An Adaptive Neurosymbolic Method for Causal Effect Estimation}
\author {
    Abbavaram Gowtham Reddy,
    Vineeth N Balasubramanian
}
\affiliations {
   Indian Institute of Technology Hyderabad, India\\
   
    cs19resch11002@iith.ac.in, vineethnb@iith.ac.in
}

\begin{document}

\maketitle

\begin{abstract}
Causal effect estimation from observational data is a central problem in causal inference. Methods based on potential outcomes framework solve this problem by exploiting inductive biases and heuristics from causal inference. Each of these methods addresses a specific aspect of causal effect estimation, such as controlling propensity score, enforcing randomization, etc., by designing neural network (NN) architectures and regularizers. In this paper, we propose an adaptive method called Neurosymbolic Causal Effect Estimator (NESTER), a generalized method for causal effect estimation. NESTER integrates the ideas used in existing methods based on multi-head NNs for causal effect estimation into one framework. We design a Domain Specific Language (DSL) tailored for causal effect estimation based on causal inductive biases used in literature. We conduct a theoretical analysis to investigate NESTER's efficacy in estimating causal effects. Our comprehensive empirical results show that NESTER performs better than state-of-the-art methods on benchmark datasets. 
% Code is available at \url{https://github.com/gautam0707/NESTER}.
\end{abstract}

\section{Introduction}
\label{sec introduction}

Causal effect estimation measures the effect of a treatment variable on an outcome variable (e.g., the effect of a medicine on recovery). Randomized Controlled Trials (RCTs), where individuals are randomly split into  \textit{treated} and \textit{control} groups, are considered the gold standard approach for causal effect estimation~\cite{rct}. However, RCTs are often: (i) unethical (e.g., to find the causal effect of smoking on lung disease, a randomly chosen person cannot be forced to smoke), and/or (ii) impossible/infeasible (e.g., in finding the causal effect of blood pressure on the risk of an adverse cardiac event, it is impossible to intervene on the same patient with and without high blood pressure with all other parameters the same)~\cite{rct_cost,carey2016some}. These limitations leave us with observational data to compute causal effects.

Observational data, similar to RCTs, suffer from \textit{the fundamental problem of causal inference}~\cite{pearl2009causality}, \textit{viz.} for any individual, we cannot observe all potential outcomes at the same time (e.g., we cannot uniquely record the same person's medical condition at a given time to two different treatments individually, say, on consuming a medicinal drug and an alternate placebo). Observational data also suffers from \textit{selection bias} (e.g., certain age groups are more likely to take certain kinds of medication compared to other age groups)~\cite{collier1996insights}. For these reasons, estimating unbiased causal effects from observational data is challenging. However, due to the many use cases in the real-world, estimating causal effects from observational data has remained an important problem in causal inference~\cite{rosenbaum1983central,rosenbaum1985constructing,neyman-rubin,morgan_winship_2014}, with recent efforts leveraging learning-based methods to this end~\cite{curth2021nonparametric}.

Simpson's paradox~\cite{pearl2016causal} underpins the necessity of choosing the correct set of features to \textit{control/adjust} for estimating causal effects from observational data. The Pearlian framework~\cite{pearl2009causality} uses graphical criteria such as \textit{back-door} criterion and \textit{front-door} criterion depending on the available adjustment variables and identifiability conditions. However, the Pearlian framework requires knowledge of the underlying causal graph, which is not feasible for many real-world scenarios. On the other hand, under the \textit{ignorability} assumption, methods based on the \textit{Rubin-Neyman potential outcomes framework}~\cite{po}, assume that a known set of observed features to control is available. However, as discussed above, observational data suffers from issues such as selection bias, leading to biased estimates of causal effects. Various methods have been proposed to address one or more of these issues in recent literature~\cite{tarnet,dragon_net,curth2021nonparametric}.

\begin{figure}%[28]{r}{0.55\textwidth}
    \centering 
    \vspace{5pt}
    \scalebox{0.53}{
    \tikzset{every picture/.style={line width=0.75pt}} %set default line width to 0.75pt  
    \input{images/tarnet.tikz}
}
    \caption{(a) TARNet architecture. $\mathbf{X}$ is feature vector, $t$ is treatment, $\hat{Y}_1, \hat{Y}_0$ are estimated potential outcomes and $\phi$ is learned representation at the end of shared layers. (b) Program $\mathcal{P}_T$ synthesized by NESTER using our Domain-Specific Language (DSL) (Tab~\ref{tab:dsl_causal_inference}) that is functionally similar to TARNet. Colors are used to show the equivalence between the components of TARNet and $\mathcal{P}_T$.}
\label{fig:comparison}
\end{figure}
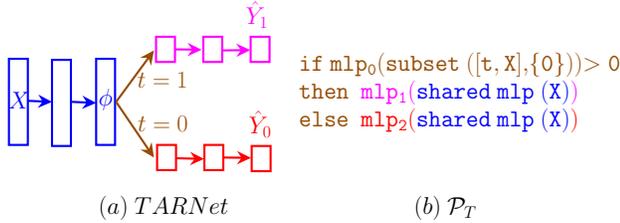

This paper provides a pathway to integrate existing solutions based on the potential outcomes framework, especially the methods based on multi-head neural network (NN) architectures with regularizers, into a single framework. We propose an adaptive method called \underline{NE}uro\underline{S}ymbolic causal effec\underline{T} \underline{E}stimato\underline{R} (NESTER) that automatically synthesizes a neurosymbolic program for estimating causal effects. Synthesized programs by NESTER can instantiate existing methods based on multi-head NN architectures with regularizers as special cases. For example, as shown in Fig~\ref{fig:comparison}, NESTER can synthesize a program $\mathcal{P}_T$ that is functionally similar to that of TARNet architecture~\cite{tarnet}. The two NN heads of TARNet corresponding to the two treatment values $t=1, t=0$ can be seen as implementing an $\mathtt{if-then-else}$ program primitive (see \S~\ref{sec_method} for detailed explanation). In $\mathcal{P}_T$, if `$\mathtt{mlp_0(subset([t,\mathbf{X}], \{0\}))>0}$', `$\mathtt{then}$' clause gets executed otherwise `$\mathtt{else}$' clause gets executed. Subscripts $1,2$ in $\mathtt{mlp_1}, \mathtt{mlp_2}$ indicate two different instantiations of the primitive $\mathtt{mlp}$ (acronym for multi-layer perceptron) corresponding to the two heads of TARNet. $\mathtt{shared\ mlp}$ primitives, without subscripts, indicate a common primitive whose output is shared by $\mathtt{mlp_1}, \mathtt{mlp_2}$ similar to how two heads of TARNet share the representation $\phi$.

As part of our proposed method NESTER, we design a Domain-Specific Language (DSL) (a specific \textit{context-free grammar}) of program primitives such as $\mathtt{if-then-else, mlp}$ containing learnable components, which is then used by neurosymbolic program synthesis (NPS) module to synthesize programs. This process is equivalent to assembling the primitives in our DSL to obtain a model architecture/workflow. In other words, NESTER learns to adaptively synthesize differentiable programs for a given set of input-output examples, wherein the sequence of learnable program modules provides an overall network architecture used to estimate causal effects.

NPS methods synthesize programs using a DSL of program primitives that satisfy given observational data of input-output pairs so that the synthesized programs generalize well to unseen inputs (see Appendix \S~\ref{flashfill},~\ref{xor} for examples)~\cite{Biermmann,gulwani2011automating,parisotto2016neuro}. Recently, various NN-based techniques have been proposed to perform NPS~\cite{parisotto2016neuro,houdini,gaunt17a,bosnjak17a,crossbeam,percepreason,near,dpads}. We use an NPS paradigm where each program primitive is a differentiable module~\cite{near,dpads}. Such \textit{differentiable programs} simultaneously optimize program primitive parameters while learning the overall program structure. In this paper, to synthesize programs, we use (i) \textit{neural admissible relaxation} (NEAR)~\cite{near}, which uses NNs as relaxations of partial programs while searching the program space and (ii) \textit{domain-specific program architecture differentiable synthesis} (dPads)~\cite{dpads} that improves on time complexity by learning the probability distribution of program architectures in a continuous relaxation of the search space of DSL grammar rules. Our key contributions are summarized below.

\begin{itemize}[leftmargin=*]
\setlength \itemsep{-0.2em}
    \item We develop an adaptive neurosymbolic method that can learn to synthesize programs for estimating causal effects. Such a method is not restricted by its architecture and is easy to implement and extend. To the best of our knowledge, this is the first neurosymbolic approach to estimate causal effects.
    \item We propose a domain-specific language (DSL) for causal effect estimation, whose program primitives are inspired by causal effect estimation efforts in literature.
    \item We theoretically study the universal approximation ability of a synthesized neurosymbolic program and show how this provides a pathway for our method for causal effect estimation. We show that the proposed method can be viewed as a generalization of causal effect estimation methods based on multi-head NN architectures.
    \item We perform comprehensive empirical studies on multiple benchmark datasets where NESTER outperforms existing state-of-the-art models.
\end{itemize}

\section{Related Work}
\label{sec related work}
\noindent \textbf{Matching and Covariate Adjustment Methods:}
Early methods for causal effect estimation are primarily based on matching techniques~\cite{neyman-rubin, morgan_winship_2014} where similar samples in treatment and control groups are compared using methods such as nearest neighbor matching~\cite{stuart2010matching} and propensity score matching~\cite{rosenbaum1983central}. For example, in nearest neighbor matching, for each sample in the treatment group, the nearest point from the control group w.r.t. Euclidean distance is identified, and the difference in observed outcomes between the treatment and corresponding control group samples is the estimate of causal effect. However, such matching techniques do not scale to high-dimensional data~\cite{matching}. Another family of methods estimates causal effects using covariate adjustment~\cite{pearl2009causality}. Assuming the availability of a sufficient adjustment set, these models fit conditional probabilities given the treatment variable and a sufficient adjustment set. Such models are, however, known to suffer from high variance in the estimated causal effects~\cite{tarnet}. Covariate balancing through weighting is another technique to control the confounding bias in estimating causal effects~\cite{rosenbaum1983central,overlap,genetic,matchingon}. As noted in~\cite{balancing}, such methods face challenges with large weights and high-dimensional inputs. Besides, leveraging the success of learning-based methods has yielded significantly better performance in recent years. 

\noindent \textbf{Representation Learning-based Methods:}
Recent methods to estimate causal effects are largely been based on multi-head NN architectures (NN architectures which branch out into different heads for different treatments) equipped with regularizers~\cite{tarnet,dragon_net,curth2021on,dr_net,chu_balanced}. CFR~\cite{tarnet} is a two-headed NN architecture with a regularizer that forces latent representations of treatment and control groups close to each other to adjust confounding features. Dragonnet~\cite{dragon_net} is a two-headed NN architecture with a regularizer that predicts treatment value from latent representations; this allows pre-treatment covariates to be used in predicting potential outcomes. Considering multiple treatment values and continuous dosage for each treatment, \cite{dr_net} devised an NN architecture with multiple heads for multiple treatments, and multiple sub-heads from each of the treatment-specific heads to model (discretized) dosage values. Assuming that potential outcomes are strongly related,~\cite{curth2021nonparametric,curth2021on} proposed techniques that improve existing models using the structural similarities between potential outcomes. These methods, however, have a fixed architecture design, and each addresses a specific problem in estimating causal effects. Our approach is also NN-based but uses a neurosymbolic approach to automatically synthesize an architecture, allowing it to generate different programs for different observational data. Causal discovery-based effect estimation and generative modeling-based effect estimation are discussed in Appendix \S~\ref{sec additional related work} along with the differences between NPS and neural architecture search (NAS).

\noindent \textbf{Neurosymbolic Program Synthesis (NPS):} Program synthesis, \textit{viz.} automatically learning a program that satisfies given input-output pairs~\cite{Biermmann}, is helpful in diverse tasks such as low-level bit manipulation code~\cite{solar}, data structure manipulations~\cite{Lezama}, and regular expression-based string generation~\cite{gulwani2011automating}. For each task, a specific DSL is used to synthesize programs. Even with a small DSL, the number of programs that can be synthesized is very large. Several techniques such as greedy enumeration, Monte Carlo sampling, Monte Carlo tree search~\cite{kocsis2006bandit}, evolutionary algorithms~\cite{houdini}, and recently, node pruning with neural admissible relaxation (NEAR)~\cite{near} have been proposed to search for optimal programs from a vast search space efficiently. Improving NEAR, dPads~\cite{dpads} propose differentiable program architecture synthesis that avoids the problem of combinatorial search required to find optimal programs. We use both NEAR and dPads to implement our method. 
 
\section{Background and Problem Formulation}
\label{sec_background}
Let $\mathcal{D}=\{(t_i,\mathbf{x}_i,y_i)\}_{i=1}^N$ be a set of $N$ observational data points. $t_i \in \mathbb{R}$ denotes the treatment variable, $\mathbf{x}_i\in \mathbb{R}^n$ denotes the $n-$dimensional feature vector, and $y_i\in \mathbb{R}$ denotes the observed potential outcome. Each $(t_i,\mathbf{x}_i, y_i)$ is randomly sampled from $p(T,\mathbf{X},Y)$, where $T, \mathbf{X}, $ and $Y$ are the corresponding random variables. In a binary treatment setting ($t\in\{0,1\}$), for the $i^{th}$ observation, let $Y^0_i$ denote the true potential outcome under treatment $t_i=0$ and $Y^1_i$ denote the true potential outcome under treatment $t_i=1$. Because of \textit{the fundamental problem of causal inference}, we observe only one of $Y^0_i, Y^1_i$ for a given $[t_i, \mathbf{x}_i]$. Hence, $y_i$ can be expressed in terms of $Y^0_i, Y^1_i$ as $y_i =  t_i Y^1_i + (1-t_i) Y^0_i$. One of the goals in causal effect estimation from observational data is to learn an estimator $f(t,\mathbf{x})$ such that the difference in estimated potential outcomes under the treatments $t=1$ and $t=0$, $f(1,\mathbf{x}_i)-f(0,\mathbf{x}_i)$, is close to the difference in true potential outcomes: $Y^1_i-Y^0_i; \ \forall i$. This difference for a specific instance $i$ is called the \textit{Individual Causal Effect (ICE)}.
\begin{definition}
    The Individual Causal Effect (ICE) of $T$ on $Y$ for an instance $\mathbf{x}\sim \mathbf{X}$ is defined as
    \begin{equation}
    \small
    \label{eqn_ite}
    ICE_T^Y(\mathbf{x}) \coloneqq  \mathbb{E}[Y^1-Y^0|\mathbf{x}]
\end{equation}
\end{definition}
\begin{definition}
\label{def: pehe}
The expected Precision in Estimation of Heterogeneous Effect ($\epsilon_{PEHE}$) using $f(t,\mathbf{x})$ is defined as
    \begin{equation}
    \small
    \label{eqn_pehe}
    \epsilon_{PEHE} (f) \coloneqq \underset{\mathbf{x}\sim \mathbf{X}}{\mathbb{E}} \big[ ((f(1,\mathbf{x})-f(0,\mathbf{x})) - ICE_T^Y(\mathbf{x}))^2 \big]
\end{equation}
\end{definition}

Extending \textit{ICE} to an entire population, our goal is to estimate the \textit{Average Causal Effect (ACE)}~\cite{pearl2009causality} of the treatment variable $T$ on the outcome variable $Y$.
\begin{definition}
\label{def:ace}
    The Average Causal Effect (ACE) of $T$ on $Y$ is defined as
    \begin{equation}
    \small
    \label{eqn_ate}
    ACE_T^Y \coloneqq  \mathbb{E}[Y^1]-\mathbb{E}[Y^0]
\end{equation}
\end{definition}

\begin{definition}
\label{def eate}
    The error in estimation of Average Causal Effect ($\epsilon_{ACE}$) using $f(t,\mathbf{x})$ is defined as
    \begin{equation}
    \small
    \label{eqn_pate}
    \epsilon_{ACE}(f) \coloneqq  |\underset{\mathbf{x}\sim \mathbf{X}}{\mathbb{E}}[f(1,\mathbf{x})-f(0,\mathbf{x})] - ACE_T^Y|
\end{equation}
\end{definition}

$\mathbb{E}[Y^t]$ refers to the expected value of $Y$ when every instance in the population is given the treatment $t$ (if $t$ is not binary, causal effects are calculated w.r.t. a baseline treatment value $t^*$ i.e., $1,0$ in Defn~\ref{def:ace} are replaced with $t, t^*$ respectively). To estimate the quantities in Eqns~\ref{eqn_ite}-\ref{eqn_pate} from observational data, it is required to guarantee \textit{identifiability}. Following~\cite{tarnet,lechner2001identification, imbens2000role,dr_net,disent}, we make the following assumptions sufficient to guarantee the \textit{identifiability} of causal effects. 
% \begin{assumption}
% \label{assumption1}
% {\textbf{(Ignorability, Positivity, Stable Unit Treatment Value Assumption (SUTVA))}} Ignorability says that for a given set of pre-treatment covariates, treatment is randomly assigned, i.e., conditioned on a set of pre-treatment covariates $\mathbf{X}$, $T$ is independent of $Y^0, Y^1$ i.e., $((Y^0, Y^1) \ind T|\mathbf{X})$. \textit{Ignorability} is also referred to as \textit{no-latent-confounding} assumption. \textit{Positivity} entails that treatment assignment for each instance is not deterministic, and it must be possible to assign all treatments to each instance, i.e., $0<p(t|\mathbf{x})<1\ \ \forall t,\mathbf{x}$. \textit{SUTVA} states that the observed outcome of any instance under a treatment must be independent of the treatment assignment to other individuals.
% \end{assumption}
\begin{assumption}
\label{assumption1}
{\textbf{(i) Ignorability} \textit{(No-latent-confounding)}:} For a given set of pre-treatment covariates, treatment is randomly assigned, i.e., conditioned on a set of pre-treatment covariates $\mathbf{X}$, $T$ is independent of $Y^0, Y^1$ i.e., $((Y^0, Y^1) \ind T|\mathbf{X})$. {\textbf{(ii) Positivity}:} Treatment assignment for each instance is not deterministic, and it must be possible to assign all treatments to each instance, i.e., $0<p(t|\mathbf{x})<1\ \ \forall t,\mathbf{x}$. {\textbf{(iii) Stable Unit Treatment Value Assumption (SUTVA))}:} The observed outcome of any instance under a treatment must be independent of the treatment assignment to other individuals.
\end{assumption}
Assuming the feature vector $\mathbf{X}$ satisfies the properties in Assumptions~\ref{assumption1}, we can write $\mathbb{E}[Y^t] = \mathbb{E}_{\mathbf{x}\sim \mathbf{X}}\left[\mathbb{E}[Y|T=t, \mathbf{X}=\mathbf{x}]\right]$, called the \textit{adjustment formula}~\cite{pearl2009causality}. Using this, a simple technique to estimate $\mathbb{E}[Y|T=t, \mathbf{X}=\mathbf{x}]$ is to fit a regression model for $Y$ given $T$ and $\mathbf{X}$. Estimation of $\mathbb{E}[Y|T=t, \mathbf{X}=\mathbf{x}]$ is the primary task of many causal effect estimation methods. In the same vein, we also aim to synthesize programs that compute the quantity $\mathbb{E}[Y|T=t, \mathbf{X}=\mathbf{x}]$. Once $\mathbb{E}[Y|T=t, \mathbf{X}=\mathbf{x}]$ is estimated, the $ATE_T^Y$ can be estimated as $ATE_T^Y = \mathbb{E}_{\mathbf{x}\sim \mathbf{X}}[\mathbb{E}[Y|T=1, \mathbf{X}=\mathbf{x}]]-\mathbb{E}_{\mathbf{x}\sim \mathbf{X}}[\mathbb{E}[Y|T=0, \mathbf{X}=\mathbf{x}]]$. 

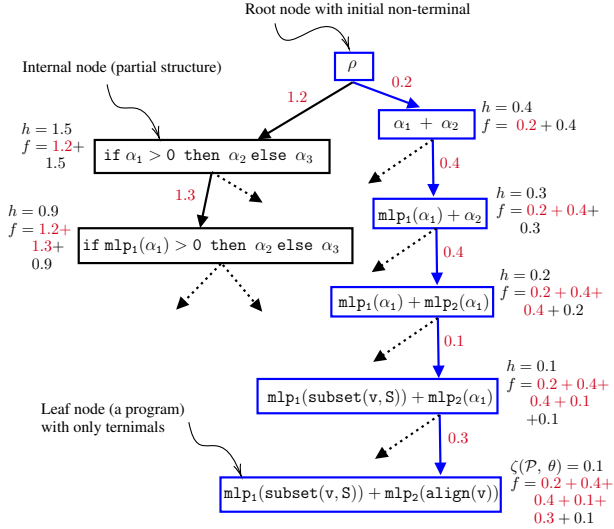
\begin{figure}
\centering
\scalebox{0.55}{
\tikzset{every picture/.style={line width=0.75pt}}
\input{images/example1.tikz}
}
\caption{Example program tree generated using DSL in Tab~\ref{tab:dsl_causal_inference}. Structural costs are shown in red color (e.g., $s(r)=0.2$ for the rule $r: \rho \rightarrow \alpha_1+\alpha_2$). $h$ is the heuristic value, and $f$ is the sum of structural cost and heuristic value. The path from the root node to a leaf node returned by $A^*$ algorithm is shown in blue color.}
\label{fig:example}
\end{figure}

\noindent \textbf{Neurosymbolic Program Synthesis (NPS):} Let $(\mathcal{P}, \theta)$ be a neurosymbolic \textit{program} where $\mathcal{P}$ denotes the program \textit{structure} and $\theta$ denotes the program \textit{parameters}. $(\mathcal{P},\theta)$ is differentiable in $\theta$. $\mathcal{P}$ is synthesized using a Context-Free Grammar (CFG)~\cite{ullman} (which is a DSL in this work). A CFG consists of a set of \textit{rules} of the form $\rho \rightarrow \alpha_1,\dots,\alpha_n$ where $\rho$ is a \textit{non-terminal} and $\alpha_1,\dots,\alpha_n$ are either \textit{non-terminals} or \textit{terminals}. A non-terminal denotes a missing sub-expression in a program structure and a terminal is a symbol that appears in a final program structure. 
Program synthesis starts with an initial non-terminal, then iteratively applies the rules to produce a series of \textit{partial structures}, \textit{viz.} structures made from one or more non-terminals and zero or more terminals. These partial structures form internal nodes of a \textit{program tree}, and the rules form the (directed) edges connecting these nodes (e.g., a rule $r$ is considered as an edge from node $u$ to node $v$ when $v$ is obtained from $u$ by applying $r$). The process continues until no non-terminals are left, i.e., we have synthesized a program. The resultant program tree's leaf nodes (a.k.a. goal nodes) contain structures consisting of only terminals (see Fig~\ref{fig:example}). 
Let $s(r)$ be the cost incurred in using the rule $r$ for generating a program structure (leaf node) or partial structure (internal node) from a given partial structure. The structural cost of any node $u$ is the sum of the structural costs of rules used to get $u$ from the root node. The structural cost $s(\mathcal{P})$ of a program $\mathcal{P}$ is defined as $s(\mathcal{P})= \sum_{r\in R(\mathcal{P})} s(r)$, where $R(\mathcal{P})$ is the multiset of rules used to create the structure $\mathcal{P}$. In this paper, we set $s(r)$ to a constant real number for all rules (e.g., $s(r)=1\hspace{0.2cm} \forall r \in R(\mathcal{P})$).
The program learning problem is usually formulated as a node search problem, i.e., starting with an empty tree, the tree is expanded by creating new partial structures and structures. 

We use $A^*$ algorithm to generate a program tree. While generating a program tree, a node $u$ with minimum $f(u)$ value is expanded next, where $f(u)=s(\mathcal{P}(u))+h(u)$ is the sum of the structural cost $s(\mathcal{P}(u))$ of the partial structure $\mathcal{P}(u)$ in $u$ and the heuristic value $h(u)$ at the node $u$. $h(u)$ underestimates the cost to reach goal node from $u$ (see Fig~\ref{fig:example} and \S~\ref{subsec_nester_algo}). While searching for an optimal program, the program parameters (and program structures) are updated simultaneously along with the synthesis of the programs. Since our goal is to synthesize a program $(\mathcal{P}, \theta)$ that estimates the quantity $\mathbb{E}[Y|T=t, \mathbf{X}=\mathbf{x}]$, which can be modeled as a regression problem, the squared error is a good choice for assessing the performance of the program $(\mathcal{P}, \theta)$ in estimating potential outcomes. Hence, for a synthesized program $(\mathcal{P},\theta)$, we define $\zeta(\mathcal{P}, \theta) = \mathbb{E}_{(t,\mathbf{x},y)\sim \mathcal{D}}[((\mathcal{P},\theta)(t,\mathbf{x})- y)^2]$ as the loss incurred by $(\mathcal{P},\theta)$ in estimating potential outcomes. The overall goal of NPS is then to find a structurally simple program with low prediction error, i.e., to solve the following optimization problem.
\begin{equation}
\small
\label{objective}
    (\mathcal{P}^*, \theta^*) = \argmin_{(\mathcal{P}, \theta)} (s(\mathcal{P}) + \zeta(\mathcal{P}, \theta))
\end{equation}

\section{NESTER: Methodology}
\label{sec_method}

The key idea of our methodology is to design a Domain-Specific Language (DSL) for causal effect estimation and subsequently leverage well-known search algorithms such as $A^*$ to synthesize programs for given observational data. We begin by discussing the proposed DSL and its connections to existing literature, followed by our overall algorithm that uses this DSL to synthesize programs.

\subsection{DSL for Causal Effect Estimation}
\label{sec:dsl}
We pose the causal effect estimation problem as the problem of mapping a set of observational input data points to the corresponding observed outcomes. Formally, given $\mathcal{D}=\{(t_i,\mathbf{x}_i,y_i)\}_{i=1}^N$, the set $\{(t_i,\mathbf{x}_i)\}_{i=1}^N$ contains inputs and the set $\{y_i\}_{i=1}^N$ contains outputs. For simplicity, let $\mathbf{v}_i = [t_i;\mathbf{x}_i]$ (concatenation of $t_i$ and $\mathbf{x}_i$) denote the $i^{th}$ input. By learning a mapping between given input-output examples, a synthesized program learns to estimate the potential outcomes for unseen inputs. To this end, we propose the following program primitives $1-7$ (basic building blocks of a synthesized program), which are differentiable and encode specific causal inductive biases in an NN model. These primitives comprise our proposed DSL, shown in Tab~\ref{tab:dsl_causal_inference}.

\begin{table}[H]
    \centering
    \scalebox{0.92}{
    \fbox{\begin{minipage}{25em}
    \begin{tabular}{l}
    $\mathtt{\mathbf{\rho}\ \mathbf{\rightarrow} if\ \mathtt{\alpha_1}>0\ then\ \mathtt{\alpha_2} \ else\ \mathtt{\alpha_3} \ |\ mlp(\alpha)\ |\ shared\ mlp(\alpha)}$\\
    $\mathtt{|\ subset(\mathbf{v}, \mathbf{S})\ |\ propensity(\mathbf{v})\ |\ align(\mathbf{v}) \ |\ \odot(\alpha_1,\alpha_2)}$\\
\\
    $\mathtt{\alpha/ \alpha_1/ \alpha_2/ \alpha_3 \rightarrow if\ \mathtt{\alpha_1}>0\ then\ \mathtt{\alpha_2} \ else\ \mathtt{\alpha_3} \ |\ mlp(\alpha)}$\\
    $\mathtt{|\ shared\ mlp(\alpha)\ |\ subset(\mathbf{v}, \mathbf{S})\ |\ propensity(\mathbf{v})}$\\
    $\mathtt{ |\ align(\mathbf{v}) \ |\ \odot(\alpha_1,\alpha_2)|\ \mathbf{v}\ |\ \mathbf{x}}$\\
    \end{tabular}
    \end{minipage}}
    }
    \caption{A DSL for the causal effect estimation in Backus-Naur form~\cite{backusnaur} and its semantics. $\rho$ is the initial non-terminal.}
    \label{tab:dsl_causal_inference}

\end{table}

\begin{enumerate}
    \item The primitive ``$\mathtt{if\ \alpha_1>0\ then\ \alpha_2\ else\ \alpha_3}$'' works similar to the equivalent programming construct. To avoid discontinuities and enable backpropagation, following~\cite{near}, we implement a smooth approximation i.e., $\mathtt{if\ \alpha_1>0\ then\ \alpha_2 \ else\ \alpha_3}$ can be written as $\mathtt{sig(\beta \cdot \alpha_1)\cdot \alpha_2 + (1-sig(\beta\cdot \alpha_1))\cdot \alpha_3}$, where $\mathtt{sig(\cdot)}$ is the \textit{sigmoid} function and $\mathtt{\beta}$ is a temperature parameter. As $\mathtt{\beta}\rightarrow 0$, the approximation approaches the usual $\mathtt{if-then-else}$. Since we implement a smooth approximation of $\mathtt{if-then-else}$, $\alpha_1$ doesn't need to be a boolean value. $\alpha_1,\alpha_2,\alpha_3$ are real numbers.
    \item The primitive ``$\mathtt{mlp(\alpha)}$'' is a multi-layer perceptron that takes a vector $\alpha$ as input and returns a real number.
    \item The primitive ``$\mathtt{shared\ mlp(\alpha)}$'' is a multi-layer perceptron. All instances of $\mathtt{shared\ mlp(\alpha)}$ in a synthesized program share the same set of parameters. It takes a vector $\alpha$ as input and returns a vector $\phi(\alpha)$ as output.
    \item The primitive ``$\mathtt{subset(\mathbf{v}, S)}$'' selects a set of features of $\mathtt{\mathbf{v}}\in \mathbb{R}^{n+1}$ indexed by the set $\mathtt{S}$ of indices. Other features of $\mathtt{\mathbf{v}}$ that are not indexed by the set $S$ are set to 0. Hence, both the input and output are vectors from $\mathbb{R}^{n+1}$. 
    \item  The primitive ``$\mathtt{propensity(\mathbf{v})}$'' works similar to ``$\mathtt{shared\ mlp}$'', taking $\mathbf{v}\in \mathbb{R}^{n+1}$ as input and returning $\phi(\mathbf{x})\in \mathbb{R}^n$ as output. In addition, during training, for a given batch of inputs, it creates a loss value $\mathcal{L}_{prop}$ that computes the binary cross-entropy loss in predicting $t$ from $\phi(\mathbf{x})$. $\mathcal{L}_{prop}$ is added to the overall loss in Eqn~\ref{objective} if $\mathtt{propensity(\mathbf{v})}$ is part of a synthesized program.
    \item  The primitive ``$\mathtt{align(\mathbf{v})}$'' works similar to ``$\mathtt{shared\ mlp}$'', taking $\mathbf{v}\in \mathbb{R}^{n+1}$ as input and returning $\phi(\mathbf{x})\in \mathbb{R}^n$ as output. In addition, during training, for a given batch of inputs, it creates a loss value $\mathcal{L}_{align}$ that computes the Maximum Mean Discrepancy (MMD) between $p(\phi(\mathbf{x})|t=1)$ and $p(\phi(\mathbf{x})|t=0)$. $\mathcal{L}_{align}$ is added to the overall loss in Eqn~\ref{objective} if $\mathtt{align(\mathbf{v})}$ is part of a synthesized program.
    \item ``$\mathtt{\odot(\alpha_1, \alpha_2)}$'' where $\odot \in \{+,\times\}$ gives additional flexibility to the program synthesizer to combine other primitives. $\mathtt{\odot(\alpha_1, \alpha_2)}$ takes two real numbers as inputs and returns a real number as output after performing the operation $\mathtt{\odot}$. 
\end{enumerate}
Note that the primitives $1-7$ are combined when the corresponding input and output dimensions match. For example, $\alpha_1$ in ``$\mathtt{if\ \alpha_1>0\ then\ \alpha_2\ else\ \alpha_3}$'' can be $\mathtt{mlp(\alpha)}$ but not $\mathtt{shared\ mlp(\alpha)}$ as the output of $\mathtt{mlp(\alpha)}$ is a real number matching the required dimension of $\alpha_1$.

\subsection{DSL Encodes Causal Inductive Biases}
\label{connections to existing methods}

As discussed in \S~\ref{sec related work}, existing learning-based causal effect estimation methods introduce inductive biases into machine learning models through regularizers or through changes in NN architectures. We now explain how the primitives of our DSL encode causal inductive biases introduced in popular causal effect estimation methods such as TARNet, CFR, Dragonnet, SNet, etc.

\noindent \textbf{Connection to multi-head NNs:}
Recall that, from \S~\ref{sec_background}, our goal is to estimate the quantity $\mathbb{E}[Y|T=t,\mathbf{X}=\mathbf{x}]$. If a single model is used to estimate both $\mathbb{E}[Y|T=1,\mathbf{X}=\mathbf{x}]$ and $\mathbb{E}[Y|T=0, \mathbf{X}=\mathbf{x}]$, it is often the case that $\mathbf{X}$ is high-dimensional and the treatment $T$ is a relatively much smaller set of variables (often, just one variable) when compared to $\mathbf{X}$. Hence, $T$ may not impact the model when making predictions, resulting in the estimated causal effect being biased towards zero~\cite{xlearner}. Using two different models to estimate $\mathbb{E}[Y|T=1,\mathbf{X}=\mathbf{x}]$ and $\mathbb{E}[Y|T=0, \mathbf{X}=\mathbf{x}]$ suffers from high variance in estimating causal effect due to limited data in treatment-specific sub-groups as well as from selection bias.

To mitigate the aforementioned issues, \cite{tarnet} propose an NN architecture in which two separate heads for treatment-specific outcomes are spawned from a shared representation layer. In NPS, to implement a shared representation layer, we can leverage the $\mathtt{shared\ mlp(\alpha)}$ primitive as explained in Fig~\ref{fig:comparison}. To implement the two-heads spanning from latent representation, an NPS can leverage the $\mathtt{if-then-else, mlp}$ and $\mathtt{subset}$ primitives. That is, in ``$\mathtt{if\ \alpha_1>0\ then\ \alpha_2\ else\ \alpha_3}$'', we can substitute ``$\mathtt{mlp(subset(\mathbf{v}, \{0\}))}$'' in place of $\alpha_1$ for conditional check and then accordingly execute $\mathtt{\alpha_2}$ or $\mathtt{\alpha_3}$ (a few points to recall at this juncture are: (i) $\mathtt{mlp}$ returns a real number, (ii) $t$ is at index $0$ in $\mathtt{\mathbf{v}}$, (iii) due to smooth approximation of $\mathtt{if-then-else}$, $\alpha_1$ doesn't need to evaluate to a boolean value). An NN with multiple heads would be implemented similarly using nested $\mathtt{if-then-else}$ primitives.
 
\noindent \textbf{Connection to propensity matching:} Recall the adjustment formula: $\mathbb{E}[Y^t] = \mathbb{E}_{\mathbf{x}\sim \mathbf{X}}\left[\mathbb{E}[Y|T=t, \mathbf{X}=\mathbf{x}]\right]$ where the set of features $\mathbf{X}$ are controlled. Instead of controlling the entire set $\mathbf{X}$, which may be high dimensional, it is enough to control the propensity score $p(T|\mathbf{X})$~\cite{rosenbaum1983central}. To achieve propensity matching, we employ the primitive $\mathtt{propensity(\mathbf{v})}$. Following~\cite{dragon_net}, for a given batch of data points, apart from producing a representation $\phi(\mathbf{x})$, $\mathtt{propensity(\mathbf{v})}$ creates a loss value in predicting the treatment from the learned representation $\phi(\mathbf{x})$. On the other hand, if we know which subset of features from $\mathbf{X}$ to control, we can adjust only those subset of features. If we are unsure of the specific features to control, multiple instances of $\mathtt{subset(\mathbf{v}, S)}$ with different $\mathtt{S}$ values can be used, allowing the NPS to select the appropriate subset to control for a given data.

\noindent \textbf{Connection to randomized controlled trials:} 
To improve the results from two-head NN models, CFR~\cite{tarnet} uses \textit{IPM regularization} (using Maximum Mean Discrepancy~\cite{gretton12} or Wasserstein distance~\cite{cuturi14}) on a latent layer representation before spanning two treatment-specific heads. Minimizing the IPM between treatment and control subgroups: $p(\phi(\mathbf{X})|T=1), p(\phi(\mathbf{X})|T=0)$ ensures $T\ind \phi(\mathbf{X})$ and thus mimicking randomized controlled trials~\cite{tarnet}.  The primitive $\mathtt{align(\mathbf{v})}$ is intended to achieve a similar purpose in our DSL. That is, for a given batch of data points, apart from producing a representation $\phi(\mathbf{X})$, $\mathtt{align(\mathbf{v})}$ creates a loss value equal to the MMD between $p(\phi(\mathbf{X})|t=1)$ and $p(\phi(\mathbf{X})|t=1)$.

We reiterate that we do not hard-code/pre-define the network architecture; the NPS learns to generate programs that compose primitives suitably to minimize overall loss during training. See Appendix \S~\ref{proofsofpropositions}, where we show how existing NN architectures CFR and Dragonnet are special cases of NESTER. We now present the algorithm to synthesize neurosymbolic programs for the estimation of causal effects.

\subsection{NESTER Algorithm}
\label{subsec_nester_algo}

We refer to \S~\ref{sec_background} for the background on NPS, which we build on here. We use the $A^*$ algorithm to implement NESTER. At any internal node $u$, $A^*$ algorithm relies on a \textit{heuristic value} $h(u)$ that underestimates the cost to reach the goal node from $u$. $h(u)$ decides which node to expand next. During program tree generation, non-terminals in an internal node $u$ are substituted by a type-correct NN (e.g., if the primitive $\alpha_1+\alpha_2$ returns a real number as output, the MLPs substituted for $\alpha_1, \alpha_2$ must also return real numbers as output). The training loss of the resultant program $(\mathcal{P}(u),\theta(u))$ on $\mathcal{D}$ then acts as the heuristic value $h(u)$ at node $u$~\cite{near}. We run the $A^{*}$ algorithm using this heuristic function to find programs that estimate causal effects. We outline our overall algorithm in Algorithm~\ref{nester} of Appendix \S~\ref{sec algorithm}. Algorithm~\ref{nester} returns the program that satisfies the objective in Eqn~\ref{objective} (additional losses can be added to objective~\ref{objective}; see primitives $5, 6$ in \S~\ref{sec:dsl}). Similar to traditional NN training, the parameters of the best program are chosen based on the cross-validation score. By keeping the overall program to only a limited depth allows us to build models that are efficiently learned and effective in practice with almost no additional time overhead compared to state-of-the-art methods.
\section{NESTER: Analysis}
\label{sec_nester_theory}
We analyze NESTER from two perspectives: (i) the capability of a program synthesized using NPS methods to perform causal effect estimation and (ii) the capabilities of our proposed DSL in relation to well-known learning-based causal effect estimation methods. For the former, we hypothesize that NPS is a viable candidate for estimating causal effects if the relationship between treatment and effect is a continuous function. To this end, we first state the notion of an $\epsilon$-admissible heuristic in Defn~\ref{defn_admissibility}, show how a synthesized program's training loss can serve as such an $\epsilon$-admissible heuristic in Lemma~\ref{prop1}, and then state our result in Propn~\ref{prop2}.

\begin{definition}
\label{defn_admissibility}
\textbf{($\epsilon$-Admissible Heuristic~\cite{harris,heuristics})} In an informed search algorithm, a heuristic function $h(u)$ that estimates the cost to reach the goal node $g$ from a node $u$ is said to be admissible if $h(u) \leq h^{*}(u), \forall u$ where $h^*(u)$ is the true cost to reach $g$ from $u$. Given an $\epsilon > 0$, $h(u)$ is said to be $\epsilon-$admissible if $h(u) \leq h^{*}(u)+\epsilon, \forall u$.
\end{definition}
\begin{restatable}[]{lemma}{admissibility}
\label{prop1}
\textbf{(Neural Admissible Relaxations~\cite{near})} In an informed search algorithm $\mathcal{A}$, given an internal node $u_i$ and a leaf node $u_l$, let the cost of the leaf edge $(u_i, u_l)$ be $s(r)+\zeta(\mathcal{P}, \theta^{*})$, where $\theta^{*} = \argmin_{\theta} \zeta(\mathcal{P}, \theta$) and $s(r)$ is the structural cost incurred by using rule $r$ to create $u_l$ from $u_i$. If an NN model $\mathcal{N}$ is used to substitute each non-terminal $\alpha$ of $u_i$ such that $\mathcal{N}$ can approximate $\alpha$ up to an arbitrary precision $\delta$ for any given $\delta$, the training loss of the program obtained is an $\epsilon-$admissible heuristic for $u_i$. 
\end{restatable}

\begin{restatable}[]{proposition}{guarantees}
\label{prop2}
\textbf{(Universal Approximation Result for NPS)} If the interventional effect of the treatment variable $T$ on the target variable $Y$ is a continuous function $g(T)$ i.e., $\mathbb{E}[Y|do(T)] =g(T)$, using any DSL $\mathcal{L}$ for synthesizing a single-hidden-layer neural network, NESTER synthesizes a program $(\mathcal{P},\theta)$ that $\epsilon-$approximates $g$ for a given $\epsilon > 0$.
\end{restatable}

All proofs are in Appendix \S~\ref{proofsofpropositions}. Our proof follows from the universal approximation theorem for NN models~\cite{hornik}, a DSL for a single-hidden-layer NN and Lemma~\ref{prop1}. The above result shows that if the relationship between treatment and effect is a continuous function, using NESTER is a viable candidate for estimating causal effects. We next discuss the capabilities of the proposed DSL w.r.t. existing methods.

% \begin{restatable}[]{proposition}{equivalence}
% \label{prop3}
% \textbf{(Generalizability of NESTER)} Using the proposed DSL, NESTER can generate programs $(\mathcal{P}_{T},\theta_{T})$, $(\mathcal{P}_{C},\theta_{C})$, and $(\mathcal{P}_{D},\theta_{D})$, whose architectures are the same as TARNet, CFR~\cite{tarnet}, and Dragonnet~\cite{dragon_net} respectively.
% \end{restatable}
\begin{table*}
    \centering
    \scalebox{0.86}{
    \begin{tabular}{lcccccc}
    \toprule
    Datasets (Metric) $\mathbf{\rightarrow}$&\multicolumn{2}{c}{IHDP ($\epsilon_{ACE} (\downarrow)$)} & \multicolumn{2}{c}{Twins ($\epsilon_{ACE} (\downarrow)$)}&\multicolumn{2}{c}{Jobs ($\epsilon_{ACT} (\downarrow)$)}\\
    \midrule
       Methods $\mathbf{\downarrow}$ &In-Sample&Out-of-Sample&In-Sample&Out-of-Sample&In-Sample&Out-of-Sample\\
    \midrule
        OLS-1 &.73$\pm$.04&.94$\pm$.05&.0038$\pm$.0025&.0069$\pm$.0056&\cellcolor{gray!50}\textbf{.01$\pm$.00}&.08$\pm$.04\\
        
        OLS-2 &.14$\pm$.01&.31$\pm$.02&.0039$\pm$.0025&.0070$\pm$.0059&\cellcolor{gray!50}\textbf{.01$\pm$.01}&.08$\pm$.03\\
        
        k-NN &.14$\pm$.01&.90$\pm$.05&.0028$\pm$.0021&.0051$\pm$.0039&.21$\pm$.01&.13$\pm$.05\\
        
        BLR&.72$\pm$.04&.93$\pm$.05&.0057$\pm$.0036&.0334$\pm$.0092&\cellcolor{gray!50}\textbf{.01$\pm$.01}&.08$\pm$.03\\
    \midrule
        BART~&.23$\pm$.01&.34$\pm$.02&.1206$\pm$.0236&.1265$\pm$.0234&.02$\pm$.00&.08$\pm$.03\\
        
        Random Forest &.73$\pm$.05&.96$\pm$.06&.0049$\pm$.0034&.0080$\pm$.0051&.03$\pm$.01&.09$\pm$.04\\
        
        Causal Forest &.18$\pm$.01&.40$\pm$.03&.0286$\pm$.0035&.0335$\pm$.0083&.03$\pm$.01&.07$\pm$.03\\
    \midrule
        BNN &.37$\pm$.03&.42$\pm$.03&.0056$\pm$.0032&.0203$\pm$.0071&.04$\pm$.01&.09$\pm$.04\\
        
        TARNet &.26$\pm$.01&.28$\pm$.01&.0108$\pm$.0017&.0151$\pm$.0018&.05$\pm$.02&.11$\pm$.04\\
        
        MHNET&.14$\pm$.13&.37$\pm$.43&.0108$\pm$.0008&.0101$\pm$.0002&.04$\pm$.01&\cellcolor{gray!50}\textbf{.06$\pm$.02}\\
        
        GANITE &.43$\pm$.05&.49$\pm$.05&.0058$\pm$.0017&.0089$\pm$.0075&\cellcolor{gray!50}\textbf{.01$\pm$.01}&\cellcolor{gray!50}\textbf{.06$\pm$.03}\\
        
        CFR$_{WASS}$ &.25$\pm$.01&.27$\pm$.01&.0112$\pm$.0016&.0284$\pm$.0032&.04$\pm$.01&.09$\pm$.03\\
        
        Dragonnet &.16$\pm$.16&.29$\pm$.31&.0057$\pm$.0003&.0150$\pm$.0003&.04$\pm$.00&.07$\pm$.00\\
        
        CMGP &.11$\pm$.10&.13$\pm$.12&.0124$\pm$.0051&.0143$\pm$.0116&.06$\pm$.06&.09$\pm$.07\\

        TNet&.20$\pm$.18&.22$\pm$.11&.0200$\pm$.0070&.0200$\pm$.0070& .03$\pm$.01&.07$\pm$.08\\
        SNet&.09$\pm$.10&.14$\pm$.12&.0040$\pm$.0030&.0040$\pm$.0030 &.03$\pm$.03&.07$\pm$.07\\
    \midrule
        \textbf{NESTER - NEAR} &\cellcolor{gray!50}
\textbf{.05$\pm$.04}&\cellcolor{gray!50}\textbf{.05$\pm$.03}&\cellcolor{gray!50}\textbf{.0034$\pm$.0005}&.0039$\pm$.0006&\cellcolor{gray!50}\textbf{.01$\pm$.00}&.08$\pm$.08\\
        \textbf{NESTER - dPads}&\cellcolor{gray!50}\textbf{.05$\pm$.01}&\cellcolor{gray!50}\textbf{.05$\pm$.02}&.0035$\pm$.0001&\cellcolor{gray!50}\textbf{.0028$\pm$.0001}&\cellcolor{gray!50}\textbf{.01$\pm$.00}&.08$\pm$.08\\
        \bottomrule
    \end{tabular}
    }
    \caption{Results on IHDP, Twins, and Jobs datasets. Lower is better. The best numbers are in bold. Simple machine learning models, ensemble models, and neural network-based models are separated using horizontal lines. OLS-1 refers to Ordinary Least Squares with treatment as a feature, OLS-2 refers to OLS with two regressors for two treatments. Further analysis on k-NN results and dataset details are in Appendix \S~\ref{experimental setup}}
    \label{tab:results}
\end{table*}

\begin{restatable}[]{proposition}{reduction}
\label{prop4}
\textbf{(Error Bounds of NESTER)} The program $(\mathcal{P}_{C},\theta_{C})$ generated by NESTER using the proposed DSL, whose architecture is the same as CFR, has the same error bounds in estimating causal effects as that of CFR.
\end{restatable}

\noindent The above theoretical results show that the models for causal effect estimation generated by NESTER can be shown to have performance bounds similar to existing methods.

\section{Experiments and Results} 
\label{sec_expts}

We perform a comprehensive suite of experiments to study the usefulness of NESTER in estimating causal effects with our proposed DSL. Our code and instructions to reproduce the results are included in the supplementary material and will be made publicly available. 

\noindent \textbf{Datasets:}
Evaluating causal effect estimation methods requires all potential outcomes to be available (Defn~\ref{def: pehe} and Defn~\ref{def eate}), which is not possible due to \textit{the fundamental problem of causal inference}. Thus, following~\cite{tarnet,yoon2018ganite,dragon_net,balance_reg}, we experiment on two semi-synthetic datasets--Twins~\cite{twins}, IHDP~\cite{ihdp_dataset}--that are derived from real-world RCTs (see Appendix \S~\ref{experimental setup} for details). For these two datasets, ground truth potential outcomes (a.k.a. counterfactual outcomes) are synthesized and available, and hence can be used to study the effectiveness of models in predicting potential outcomes. We also experiment on one real-world dataset--Jobs~\cite{jobs}--where we observe only one potential outcome. We note that we are commensurate or better than existing work on the number of datasets studied. More details of datasets are provided in Appendix\S~\ref{experimental setup}.

\noindent \textbf{Baselines:}
We compare NESTER with 16 baselines from various categories of methods as shown in Tab~\ref{tab:results}. We implement NESTER using NEAR~\cite{near} (NESTER-NEAR) that uses neural networks as relaxations of partial programs and dPads~\cite{dpads} (NESTER-dPads) that avoids combinatorial search over possible programs using a differentiable pruning strategy.

\noindent \textbf{Evaluation Metrics:}
For the experiments on IHDP and Twins datasets where we have access to both potential outcomes, following~\cite{tarnet,yoon2018ganite,dragon_net,balance_reg}, we use the evaluation metrics: $\epsilon_{ACE}$ and $\epsilon_{PEHE}$ (Defn~\ref{def: pehe} and Defn~\ref{def eate}).
$\epsilon_{ACE}$ measures the error in the estimation of the average causal effect in a population. $\epsilon_{PEHE}$ is operated on the error in the estimation of individual causal effects. For the experiments on the Jobs dataset where we observe only one potential outcome per data point, following~\cite{tarnet,yoon2018ganite,dragon_net,balance_reg}, we use the metric \textit{error in the estimation of average causal effect on the treated} ($\epsilon_{ACT}$). Definitions and more details of these metrics are provided in Appendix \S~\ref{experimental setup}. Following~\cite{tarnet,dragon_net,yoon2018ganite}, 
We report both in-sample and out-of-sample performance w.r.t. $\epsilon_{ACE}, \epsilon_{ACT}, \sqrt{\epsilon_{PEHE}}$ in our results. Unlike traditional supervised learning, in-sample performance is non-trivial in this context, since we do not observe counterfactual outcomes (all potential outcomes) during training. Additional details on the experimental setup are presented in Appendix \S~\ref{experimental setup}.

\noindent \textbf{Results:}
The results shown in Tab~\ref{tab:results} show the superior performance of NESTER over existing methods. To understand the results, we investigate the programs synthesized by NESTER. For example, the synthesized program for Jobs dataset is shown below where the synthesized program is similar to a two-head NN architecture with conditional check based on the propensity score.
\begin{equation*}
    \begin{aligned}
        &\mathtt{if\ mlp_0(propensity(\mathbf{v}))>0\ then\ mlp_1(align(\mathbf{v}))}\\
        &\mathtt{else\ mlp_2(align(\mathbf{v}))}\\
    \end{aligned}
\end{equation*}
While the existing two head NN models perform conditional check based on treatment variable, the above program synthesized for Jobs dataset is also a valid program because it aligns with causal inductive bias that propensity score can be viewed as a direct parent of treatment variable~\cite{rosenbaum1983central, dragon_net} and thus be valid for conditional check to decide on the specific NN head to execute. Program synthesis with appropriate primitives generates such valid programs that are not considered in literature. For Twins dataset, NESTER synthesized a simple program ``$\mathtt{mlp(\mathbf{v})}$'' supporting the fact that simple programs such as linear regression performs better w.r.t. both in-sample and out-of-sample data (see Tab~\ref{tab:results}). For IHDP dataset, which is a collection of 1000 simulated datasets, we observe that the program below is synthesized more often. 
\begin{equation*}
    \begin{aligned}
        &\mathtt{if\ mlp_0(subset(\mathbf{v}, \{0,\dots,|\mathbf{v}|\}))>0\ }\\
        &\mathtt{then\ mlp_1(align(\mathbf{v}))\ else\ mlp_2(align(\mathbf{v}))}\\
    \end{aligned}
\end{equation*}
The program above is similar to a two head NN architecture where the conditional check is based on the entire feature set. To permit efficient learning (and to some degree, interpretability of the learned program, as discussed in Appendix \S\ref{appx_sec_interpretability}), we limit the program depth to utmost 5 for the main experiments. In Appendix \S~\ref{ablation}, we present results with other depths, results w.r.t. $\epsilon_{PEHE}$ metric, and the run time analysis of our method compared to baselines.

\section{Limitations and Conclusions}
\label{conclusions}

We present an adaptive method for estimating causal effects using neurosymbolic program synthesis. We propose a DSL for causal effect estimation by comparing program primitives and model building blocks. The viability and suitability of this approach are theoretically demonstrated. Our approach is validated through extensive experimentation on benchmark datasets with multiple baselines, showcasing its usefulness. Enhancing the DSL with additional program primitives and addressing potential run-time issues induced thereof is a promising future direction. Relaxing the no latent confounding can be another interesting future work. This work has no known detrimental effects.

\bibliography{aaai24}

\include{supplementary}
\end{document}

%% file: images/tarnet.tikz
\begin{tikzpicture}[x=0.75pt,y=0.75pt,yscale=-1,xscale=1]
%uncomment if require: \path (0,525); %set diagram left start at 0, and has height of 525

%Shape: Rectangle [id:dp8593177697853537] 
\draw  [color={rgb, 255:red, 0; green, 0; blue, 255 }  ,draw opacity=1 ][fill={rgb, 255:red, 255; green, 255; blue, 255 }  ,fill opacity=1 ][line width=1.5]  (13,94.5) -- (31,94.5) -- (31,176.5) -- (13,176.5) -- cycle ;
%Straight Lines [id:da4504312661560037] 
\draw [color={rgb, 255:red, 0; green, 0; blue, 255 }  ,draw opacity=1 ][line width=1.5]    (31,134.5) -- (50,135.33) ;
\draw [shift={(54,135.5)}, rotate = 182.49] [fill={rgb, 255:red, 0; green, 0; blue, 255 }  ,fill opacity=1 ][line width=0.08]  [draw opacity=0] (13.4,-6.43) -- (0,0) -- (13.4,6.44) -- (8.9,0) -- cycle    ;
%Straight Lines [id:da6702295989991198] 
\draw [color={rgb, 255:red, 0; green, 0; blue, 255 }  ,draw opacity=1 ][line width=1.5]    (73,135.5) -- (92,136.33) ;
\draw [shift={(96,136.5)}, rotate = 182.49] [fill={rgb, 255:red, 0; green, 0; blue, 255 }  ,fill opacity=1 ][line width=0.08]  [draw opacity=0] (13.4,-6.43) -- (0,0) -- (13.4,6.44) -- (8.9,0) -- cycle    ;
%Shape: Rectangle [id:dp4889054461494067] 
\draw  [color={rgb, 255:red, 0; green, 0; blue, 255 }  ,draw opacity=1 ][fill={rgb, 255:red, 255; green, 255; blue, 255 }  ,fill opacity=1 ][line width=1.5]  (54,95.5) -- (72,95.5) -- (72,177.5) -- (54,177.5) -- cycle ;
%Shape: Rectangle [id:dp1995365810066867] 
\draw  [color={rgb, 255:red, 0; green, 0; blue, 255 }  ,draw opacity=1 ][fill={rgb, 255:red, 255; green, 255; blue, 255 }  ,fill opacity=1 ][line width=1.5]  (96,94.5) -- (114,94.5) -- (114,176.5) -- (96,176.5) -- cycle ;
%Shape: Rectangle [id:dp5762962934423812] 
\draw  [color={rgb, 255:red, 255; green, 0; blue, 255 }  ,draw opacity=1 ][fill={rgb, 255:red, 255; green, 255; blue, 255 }  ,fill opacity=1 ][line width=1.5]  (153,73.5) -- (171,73.5) -- (171,97.5) -- (153,97.5) -- cycle ;
%Shape: Rectangle [id:dp7414418446801542] 
\draw  [color={rgb, 255:red, 255; green, 0; blue, 0 }  ,draw opacity=1 ][fill={rgb, 255:red, 255; green, 255; blue, 255 }  ,fill opacity=1 ][line width=1.5]  (154,176.5) -- (172,176.5) -- (172,200.5) -- (154,200.5) -- cycle ;
%Straight Lines [id:da9315622401761552] 
\draw [color={rgb, 255:red, 150; green, 75; blue, 0 }  ,draw opacity=1 ][line width=1.5]    (115,135.5) -- (148.63,89.72) ;
\draw [shift={(151,86.5)}, rotate = 126.3] [fill={rgb, 255:red, 150; green, 75; blue, 0 }  ,fill opacity=1 ][line width=0.08]  [draw opacity=0] (13.4,-6.43) -- (0,0) -- (13.4,6.44) -- (8.9,0) -- cycle    ;
%Straight Lines [id:da6474181817702674] 
\draw [color={rgb, 255:red, 150; green, 75; blue, 0 }  ,draw opacity=1 ][line width=1.5]    (115,135.5) -- (148.75,185.19) ;
\draw [shift={(151,188.5)}, rotate = 235.81] [fill={rgb, 255:red, 150; green, 75; blue, 0 }  ,fill opacity=1 ][line width=0.08]  [draw opacity=0] (13.4,-6.43) -- (0,0) -- (13.4,6.44) -- (8.9,0) -- cycle    ;
%Shape: Rectangle [id:dp1356975453608953] 
\draw  [color={rgb, 255:red, 255; green, 0; blue, 255 }  ,draw opacity=1 ][fill={rgb, 255:red, 255; green, 255; blue, 255 }  ,fill opacity=1 ][line width=1.5]  (198,73.5) -- (216,73.5) -- (216,97.5) -- (198,97.5) -- cycle ;
%Straight Lines [id:da7228162457541072] 
\draw [color={rgb, 255:red, 255; green, 0; blue, 255 }  ,draw opacity=1 ][fill={rgb, 255:red, 255; green, 0; blue, 255 }  ,fill opacity=1 ][line width=1.5]    (171,85.5) -- (192.33,85.78) ;
\draw [shift={(196.33,85.83)}, rotate = 180.75] [fill={rgb, 255:red, 255; green, 0; blue, 255 }  ,fill opacity=1 ][line width=0.08]  [draw opacity=0] (13.4,-6.43) -- (0,0) -- (13.4,6.44) -- (8.9,0) -- cycle    ;
%Shape: Rectangle [id:dp5573253410501693] 
\draw  [color={rgb, 255:red, 255; green, 0; blue, 0 }  ,draw opacity=1 ][fill={rgb, 255:red, 255; green, 255; blue, 255 }  ,fill opacity=1 ][line width=1.5]  (199,176.5) -- (217,176.5) -- (217,200.5) -- (199,200.5) -- cycle ;
%Straight Lines [id:da9731513253721154] 
\draw [color={rgb, 255:red, 255; green, 0; blue, 0 }  ,draw opacity=1 ][line width=1.5]    (172.33,189.17) -- (194.67,189.73) ;
\draw [shift={(198.67,189.83)}, rotate = 181.45] [fill={rgb, 255:red, 255; green, 0; blue, 0 }  ,fill opacity=1 ][line width=0.08]  [draw opacity=0] (13.4,-6.43) -- (0,0) -- (13.4,6.44) -- (8.9,0) -- cycle    ;
%Shape: Rectangle [id:dp5945501527720359] 
\draw  [color={rgb, 255:red, 255; green, 0; blue, 255 }  ,draw opacity=1 ][fill={rgb, 255:red, 255; green, 255; blue, 255 }  ,fill opacity=1 ][line width=1.5]  (243,73.5) -- (261,73.5) -- (261,97.5) -- (243,97.5) -- cycle ;
%Straight Lines [id:da10149400864039948] 
\draw [color={rgb, 255:red, 255; green, 0; blue, 255 }  ,draw opacity=1 ][fill={rgb, 255:red, 255; green, 0; blue, 255 }  ,fill opacity=1 ][line width=1.5]    (216.67,86.5) -- (238.67,86.22) ;
\draw [shift={(242.67,86.17)}, rotate = 179.27] [fill={rgb, 255:red, 255; green, 0; blue, 255 }  ,fill opacity=1 ][line width=0.08]  [draw opacity=0] (13.4,-6.43) -- (0,0) -- (13.4,6.44) -- (8.9,0) -- cycle    ;
%Shape: Rectangle [id:dp42429467967033463] 
\draw  [color={rgb, 255:red, 255; green, 0; blue, 0 }  ,draw opacity=1 ][fill={rgb, 255:red, 255; green, 255; blue, 255 }  ,fill opacity=1 ][line width=1.5]  (244,177.5) -- (262,177.5) -- (262,201.5) -- (244,201.5) -- cycle ;
%Straight Lines [id:da4680761172114013] 
\draw [color={rgb, 255:red, 255; green, 0; blue, 0 }  ,draw opacity=1 ][line width=1.5]    (217,188.5) -- (239.33,188.78) ;
\draw [shift={(243.33,188.83)}, rotate = 180.73] [fill={rgb, 255:red, 255; green, 0; blue, 0 }  ,fill opacity=1 ][line width=0.08]  [draw opacity=0] (13.4,-6.43) -- (0,0) -- (13.4,6.44) -- (8.9,0) -- cycle    ;

% Text Node
\draw (97,223.4) node [anchor=north west][inner sep=0.75pt]  [font=\huge]  {$( a) \ TARNet$};
% Text Node
\draw (281,85.53) node [anchor=north west][inner sep=0.75pt]  [font=\huge]  {$ \begin{array}{l}
\mathtt{\textcolor[rgb]{0.59,0.29,0}{if\ mlp}\textcolor[rgb]{0.59,0.29,0}{_{0}(}\textcolor[rgb]{0.59,0.29,0}{subset\ }\textcolor[rgb]{0.59,0.29,0}{(}\textcolor[rgb]{0.59,0.29,0}{[}\textcolor[rgb]{0.59,0.29,0}{t,X}\textcolor[rgb]{0.59,0.29,0}{]}\textcolor[rgb]{0.59,0.29,0}{,}\textcolor[rgb]{0.59,0.29,0}{\{}\textcolor[rgb]{0.59,0.29,0}{0}\textcolor[rgb]{0.59,0.29,0}{\}}\textcolor[rgb]{0.59,0.29,0}{)}\textcolor[rgb]{0.59,0.29,0}{)}\textcolor[rgb]{0.59,0.29,0}{ >0}}\\
\mathtt{\textcolor[rgb]{0.59,0.29,0}{then} \ \ \textcolor[rgb]{1,0,1}{mlp}\textcolor[rgb]{1,0,1}{_{1}}\textcolor[rgb]{1,0,1}{(}\textcolor[rgb]{0,0,1}{shared\ }\textcolor[rgb]{0,0,1}{mlp\ }\textcolor[rgb]{0,0,1}{(}\textcolor[rgb]{0,0,1}{X}\textcolor[rgb]{0,0,1}{)}}\textcolor[rgb]{1,0,1}{)}\\
\mathtt{\textcolor[rgb]{0.59,0.29,0}{else} \ \ \textcolor[rgb]{1,0,0}{mlp}\textcolor[rgb]{1,0,0}{_{2}}\textcolor[rgb]{1,0,0}{(}\textcolor[rgb]{0,0,1}{shared\ mlp\ }\textcolor[rgb]{0,0,1}{(}\textcolor[rgb]{0,0,1}{X}\textcolor[rgb]{0,0,1}{)}}\textcolor[rgb]{1,0,0}{)}\\
\end{array}$};
% Text Node
\draw (397,223.4) node [anchor=north west][inner sep=0.75pt]  [font=\huge]  {$( b) \ \mathcal{P}_{T}$};
% Text Node
\draw (11,124.23) node [anchor=north west][inner sep=0.75pt]  [font=\huge,color={rgb, 255:red, 0; green, 0; blue, 255 }  ,opacity=1 ]  {$X$};
% Text Node
\draw (133.67,105.57) node [anchor=north west][inner sep=0.75pt]  [font=\huge,color={rgb, 255:red, 150; green, 75; blue, 0 }  ,opacity=1 ]  {$t=1$};
% Text Node
\draw (134,146.9) node [anchor=north west][inner sep=0.75pt]  [font=\huge,color={rgb, 255:red, 150; green, 75; blue, 0 }  ,opacity=1 ]  {$t=0$};
% Text Node
\draw (97,121.9) node [anchor=north west][inner sep=0.75pt]  [font=\huge,color={rgb, 255:red, 0; green, 0; blue, 255 }  ,opacity=1 ]  {$\phi $};
% Text Node
\draw (237,38.4) node [anchor=north west][inner sep=0.75pt]  [font=\huge,color={rgb, 255:red, 255; green, 0; blue, 255 }  ,opacity=1 ]  {$\hat{Y}_{1}$};
% Text Node
\draw (240,142.07) node [anchor=north west][inner sep=0.75pt]  [font=\huge,color={rgb, 255:red, 255; green, 0; blue, 0 }  ,opacity=1 ]  {$\hat{Y}_{0}$};

\end{tikzpicture}

%% file: images/example1.tikz
\begin{tikzpicture}[x=0.75pt,y=0.75pt,yscale=-1,xscale=1]
%uncomment if require: \path (0,541); %set diagram left start at 0, and has height of 541

%Straight Lines [id:da1313474968795949] 
\draw [line width=1.5]    (359.33,85) -- (275.43,135.44) ;
\draw [shift={(272,137.5)}, rotate = 328.99] [fill={rgb, 255:red, 0; green, 0; blue, 0 }  ][line width=0.08]  [draw opacity=0] (11.61,-5.58) -- (0,0) -- (11.61,5.58) -- cycle    ;
%Straight Lines [id:da923438386784523] 
\draw [color={rgb, 255:red, 0; green, 0; blue, 255 }  ,draw opacity=1 ][line width=1.5]    (359.33,85) -- (417.25,106.6) ;
\draw [shift={(421,108)}, rotate = 200.45] [fill={rgb, 255:red, 0; green, 0; blue, 255 }  ,fill opacity=1 ][line width=0.08]  [draw opacity=0] (11.61,-5.58) -- (0,0) -- (11.61,5.58) -- cycle    ;
%Straight Lines [id:da42239482244121407] 
\draw [line width=1.5]    (230,167.5) -- (220.15,214.92) ;
\draw [shift={(219.33,218.83)}, rotate = 281.74] [fill={rgb, 255:red, 0; green, 0; blue, 0 }  ][line width=0.08]  [draw opacity=0] (11.61,-5.58) -- (0,0) -- (11.61,5.58) -- cycle    ;
%Straight Lines [id:da03191617355842247] 
\draw [line width=1.5]  [dash pattern={on 1.69pt off 2.76pt}]  (230,167.5) -- (273.24,193.44) ;
\draw [shift={(276.67,195.5)}, rotate = 210.96] [fill={rgb, 255:red, 0; green, 0; blue, 0 }  ][line width=0.08]  [draw opacity=0] (11.61,-5.58) -- (0,0) -- (11.61,5.58) -- cycle    ;
%Straight Lines [id:da7360409316437804] 
\draw [color={rgb, 255:red, 0; green, 0; blue, 255 }  ,draw opacity=1 ][line width=1.5]    (432.67,137.67) -- (433.28,186) ;
\draw [shift={(433.33,190)}, rotate = 269.27] [fill={rgb, 255:red, 0; green, 0; blue, 255 }  ,fill opacity=1 ][line width=0.08]  [draw opacity=0] (11.61,-5.58) -- (0,0) -- (11.61,5.58) -- cycle    ;
%Straight Lines [id:da5317557106259179] 
\draw [line width=1.5]  [dash pattern={on 1.69pt off 2.76pt}]  (432.67,137.67) -- (376.64,175.75) ;
\draw [shift={(373.33,178)}, rotate = 325.79] [fill={rgb, 255:red, 0; green, 0; blue, 0 }  ][line width=0.08]  [draw opacity=0] (11.61,-5.58) -- (0,0) -- (11.61,5.58) -- cycle    ;
%Straight Lines [id:da4123522213435479] 
\draw [color={rgb, 255:red, 0; green, 0; blue, 255 }  ,draw opacity=1 ][line width=1.5]    (437.33,301.33) -- (438.11,352.67) ;
\draw [shift={(438.17,356.67)}, rotate = 269.14] [fill={rgb, 255:red, 0; green, 0; blue, 255 }  ,fill opacity=1 ][line width=0.08]  [draw opacity=0] (11.61,-5.58) -- (0,0) -- (11.61,5.58) -- cycle    ;
%Straight Lines [id:da3814949266052855] 
\draw [line width=1.5]  [dash pattern={on 1.69pt off 2.76pt}]  (238.33,251.67) -- (200.12,290.89) ;
\draw [shift={(197.33,293.75)}, rotate = 314.25] [fill={rgb, 255:red, 0; green, 0; blue, 0 }  ][line width=0.08]  [draw opacity=0] (11.61,-5.58) -- (0,0) -- (11.61,5.58) -- cycle    ;
%Straight Lines [id:da6463894922598901] 
\draw [line width=1.5]  [dash pattern={on 1.69pt off 2.76pt}]  (240.17,252.67) -- (273.76,292.61) ;
\draw [shift={(276.33,295.67)}, rotate = 229.93] [fill={rgb, 255:red, 0; green, 0; blue, 0 }  ][line width=0.08]  [draw opacity=0] (11.61,-5.58) -- (0,0) -- (11.61,5.58) -- cycle    ;
%Shape: Rectangle [id:dp4974848724967008] 
\draw  [color={rgb, 255:red, 0; green, 0; blue, 255 }  ,draw opacity=1 ][line width=1.5]  (343,58) -- (378,58) -- (378,83) -- (343,83) -- cycle ;
%Shape: Rectangle [id:dp0028480153069052605] 
\draw  [color={rgb, 255:red, 0; green, 0; blue, 255 }  ,draw opacity=1 ][line width=1.5]  (383.33,110.67) -- (471,110.67) -- (471,136.67) -- (383.33,136.67) -- cycle ;
%Shape: Rectangle [id:dp1879942863770837] 
\draw  [color={rgb, 255:red, 0; green, 0; blue, 255 }  ,draw opacity=1 ][line width=1.5]  (378.33,192) -- (482,192) -- (482,219) -- (378.33,219) -- cycle ;
%Shape: Rectangle [id:dp02920615630857515] 
\draw  [color={rgb, 255:red, 0; green, 0; blue, 255 }  ,draw opacity=1 ][line width=1.5]  (272,357.33) -- (492,357.33) -- (492,388.83) -- (272,388.83) -- cycle ;
%Shape: Rectangle [id:dp12929733687500156] 
\draw  [line width=1.5]  (122.67,138) -- (338,138) -- (338,168.33) -- (122.67,168.33) -- cycle ;
%Shape: Rectangle [id:dp20772332791344483] 
\draw  [line width=1.5]  (108,219) -- (359,219) -- (359,251) -- (108,251) -- cycle ;
%Straight Lines [id:da8336319978068009] 
\draw [color={rgb, 255:red, 0; green, 0; blue, 255 }  ,draw opacity=1 ][line width=1.5]    (436,219) -- (436.62,267.33) ;
\draw [shift={(436.67,271.33)}, rotate = 269.27] [fill={rgb, 255:red, 0; green, 0; blue, 255 }  ,fill opacity=1 ][line width=0.08]  [draw opacity=0] (11.61,-5.58) -- (0,0) -- (11.61,5.58) -- cycle    ;
%Straight Lines [id:da756487675298304] 
\draw [line width=1.5]  [dash pattern={on 1.69pt off 2.76pt}]  (436,219) -- (379.97,257.08) ;
\draw [shift={(376.67,259.33)}, rotate = 325.79] [fill={rgb, 255:red, 0; green, 0; blue, 0 }  ][line width=0.08]  [draw opacity=0] (11.61,-5.58) -- (0,0) -- (11.61,5.58) -- cycle    ;
%Shape: Rectangle [id:dp8516233788173994] 
\draw  [color={rgb, 255:red, 0; green, 0; blue, 255 }  ,draw opacity=1 ][line width=1.5]  (339.67,273) -- (488,273) -- (488,300) -- (339.67,300) -- cycle ;
%Straight Lines [id:da4035032952105687] 
\draw [color={rgb, 255:red, 0; green, 0; blue, 255 }  ,draw opacity=1 ][line width=1.5]    (439,390.33) -- (439.77,441.67) ;
\draw [shift={(439.83,445.67)}, rotate = 269.14] [fill={rgb, 255:red, 0; green, 0; blue, 255 }  ,fill opacity=1 ][line width=0.08]  [draw opacity=0] (11.61,-5.58) -- (0,0) -- (11.61,5.58) -- cycle    ;
%Shape: Rectangle [id:dp0010992824467052076] 
\draw  [color={rgb, 255:red, 0; green, 0; blue, 255 }  ,draw opacity=1 ][line width=1.5]  (238,447.67) -- (495,447.67) -- (495,479.17) -- (238,479.17) -- cycle ;
%Straight Lines [id:da8096527163017537] 
\draw [line width=1.5]  [dash pattern={on 1.69pt off 2.76pt}]  (437.33,301.33) -- (381.31,339.42) ;
\draw [shift={(378,341.67)}, rotate = 325.79] [fill={rgb, 255:red, 0; green, 0; blue, 0 }  ][line width=0.08]  [draw opacity=0] (11.61,-5.58) -- (0,0) -- (11.61,5.58) -- cycle    ;
%Straight Lines [id:da14467074288006398] 
\draw [line width=1.5]  [dash pattern={on 1.69pt off 2.76pt}]  (439,390.33) -- (382.97,428.42) ;
\draw [shift={(379.67,430.67)}, rotate = 325.79] [fill={rgb, 255:red, 0; green, 0; blue, 0 }  ][line width=0.08]  [draw opacity=0] (11.61,-5.58) -- (0,0) -- (11.61,5.58) -- cycle    ;
%Curve Lines [id:da3703572462662993] 
\draw    (134,87) .. controls (142.91,133.04) and (161.62,82.53) .. (181.4,134.88) ;
\draw [shift={(182,136.5)}, rotate = 250.02] [color={rgb, 255:red, 0; green, 0; blue, 0 }  ][line width=0.75]    (10.93,-3.29) .. controls (6.95,-1.4) and (3.31,-0.3) .. (0,0) .. controls (3.31,0.3) and (6.95,1.4) .. (10.93,3.29)   ;
%Curve Lines [id:da6872379586983893] 
\draw    (209,397) .. controls (217.91,443.04) and (236.62,392.53) .. (256.4,444.88) ;
\draw [shift={(257,446.5)}, rotate = 250.02] [color={rgb, 255:red, 0; green, 0; blue, 0 }  ][line width=0.75]    (10.93,-3.29) .. controls (6.95,-1.4) and (3.31,-0.3) .. (0,0) .. controls (3.31,0.3) and (6.95,1.4) .. (10.93,3.29)   ;
%Curve Lines [id:da7263177378632352] 
\draw    (288,31) .. controls (296.91,77.04) and (335.22,1.05) .. (355.39,52.89) ;
\draw [shift={(356,54.5)}, rotate = 250.02] [color={rgb, 255:red, 0; green, 0; blue, 0 }  ][line width=0.75]    (10.93,-3.29) .. controls (6.95,-1.4) and (3.31,-0.3) .. (0,0) .. controls (3.31,0.3) and (6.95,1.4) .. (10.93,3.29)   ;

% Text Node
\draw (353.33,63.4) node [anchor=north west][inner sep=0.75pt]  [font=\Large]  {$\rho $};
% Text Node
\draw (129.13,145.53) node [anchor=north west][inner sep=0.75pt]  [font=\Large]  {$\mathtt{if\ \alpha _{1}  >0\ \ then\ \ } \alpha _{2} \ \mathtt{else\ \ } \alpha _{3}$};
% Text Node
\draw (395.47,115.87) node [anchor=north west][inner sep=0.75pt]  [font=\Large]  {$\mathtt{\alpha _{1} \ } +\ \alpha _{2}$};
% Text Node
\draw (381.33,198.4) node [anchor=north west][inner sep=0.75pt]  [font=\Large]  {$\mathtt{mlp_{1}( \alpha _{1})} +\alpha _{2}$};
% Text Node
\draw (279.8,364.53) node [anchor=north west][inner sep=0.75pt]  [font=\Large]  {$\mathtt{mlp_{1}( subset( v,S)}) +\mathtt{mlp_{2}( \alpha _{1})}$};
% Text Node
\draw (110,225.4) node [anchor=north west][inner sep=0.75pt]  [font=\Large]  {$\mathtt{if\ mlp_{1}( \alpha _{1})  >0\ \ then\ \ } \alpha _{2} \ \mathtt{else\ \ } \alpha _{3}$};
% Text Node
\draw (47.33,119.4) node [anchor=north west][inner sep=0.75pt]  [font=\large]  {$ \begin{array}{l}
h=1.5\\
f=\textcolor[rgb]{0.82,0.01,0.11}{1.2} +\\
\ \ \ \ \ \ 1.5
\end{array}$};
% Text Node
\draw (495.67,428.73) node [anchor=north west][inner sep=0.75pt]  [font=\large]  {$ \begin{array}{l}
\zeta (\mathcal{P} ,\ \theta ) =0.1\\
f=\textcolor[rgb]{0.82,0.01,0.11}{0.2+0.4+}\\
\textcolor[rgb]{0.82,0.01,0.11}{\ \ \ \ \ \ 0.4+0.1+}\\
\textcolor[rgb]{0.82,0.01,0.11}{\ \ \ \ \ \ 0.3} +0.1
\end{array}$};
% Text Node
\draw (482.67,178.4) node [anchor=north west][inner sep=0.75pt]  [font=\large]  {$ \begin{array}{l}
h=0.3\\
f=\textcolor[rgb]{0.82,0.01,0.11}{0.2+0.4} +\\
\ \ \ \ \ \ 0.3
\end{array}$};
% Text Node
\draw (469.33,100.07) node [anchor=north west][inner sep=0.75pt]  [font=\large]  {$ \begin{array}{l}
h=0.4\\
f=\ \textcolor[rgb]{0.82,0.01,0.11}{0.2} +0.4\ 
\end{array}$};
% Text Node
\draw (35,195.4) node [anchor=north west][inner sep=0.75pt]  [font=\large]  {$ \begin{array}{l}
h=0.9\\
f=\textcolor[rgb]{0.82,0.01,0.11}{1.2+}\\
\textcolor[rgb]{0.82,0.01,0.11}{\ \ \ \ \ \ 1.3} +\\
\ \ \ \ \ \ 0.9
\end{array}$};
% Text Node
\draw (297.67,92.4) node [anchor=north west][inner sep=0.75pt]  [font=\large]  {$\textcolor[rgb]{0.82,0.01,0.11}{1.2}$};
% Text Node
\draw (393.67,78.4) node [anchor=north west][inner sep=0.75pt]  [font=\large]  {$\textcolor[rgb]{0.82,0.01,0.11}{0.2}$};
% Text Node
\draw (437.67,153.07) node [anchor=north west][inner sep=0.75pt]  [font=\large]  {$\textcolor[rgb]{0.82,0.01,0.11}{0.4}$};
% Text Node
\draw (441.67,316.4) node [anchor=north west][inner sep=0.75pt]  [font=\large]  {$\textcolor[rgb]{0.82,0.01,0.11}{0.1}$};
% Text Node
\draw (195,182.4) node [anchor=north west][inner sep=0.75pt]  [font=\large]  {$\textcolor[rgb]{0.82,0.01,0.11}{1.3}$};
% Text Node
\draw (259,11) node [anchor=north west][inner sep=0.75pt]  [font=\large] [align=left] {Root node with initial non-terminal};
% Text Node
\draw (55,64) node [anchor=north west][inner sep=0.75pt]  [font=\large] [align=left] {Internal node (partial structure)};
% Text Node
\draw (72,378.33) node [anchor=north west][inner sep=0.75pt]  [font=\large] [align=left] {Leaf node (a program) \\with only ternimals};
% Text Node
\draw (345.13,278.2) node [anchor=north west][inner sep=0.75pt]  [font=\Large]  {$\mathtt{mlp_{1}( \alpha _{1})} +\mathtt{mlp_{2}( \alpha _{1})}$};
% Text Node
\draw (486,253.73) node [anchor=north west][inner sep=0.75pt]  [font=\large]  {$ \begin{array}{l}
h=0.2\\
f=\textcolor[rgb]{0.82,0.01,0.11}{0.2+0.4+}\\
\textcolor[rgb]{0.82,0.01,0.11}{\ \ \ \ \ \ 0.4} +0.2
\end{array}$};
% Text Node
\draw (441,234.4) node [anchor=north west][inner sep=0.75pt]  [font=\large]  {$\textcolor[rgb]{0.82,0.01,0.11}{0.4}$};
% Text Node
\draw (240,453.07) node [anchor=north west][inner sep=0.75pt]  [font=\Large]  {$\mathtt{mlp_{1}( subset( v,S)}) +\mathtt{mlp_{2}( align( v))}$};
% Text Node
\draw (445.33,405.4) node [anchor=north west][inner sep=0.75pt]  [font=\large]  {$\textcolor[rgb]{0.82,0.01,0.11}{0.3}$};
% Text Node
\draw (492,337.73) node [anchor=north west][inner sep=0.75pt]  [font=\large]  {$ \begin{array}{l}
h=0.1\\
f=\textcolor[rgb]{0.82,0.01,0.11}{0.2+0.4+}\\
\textcolor[rgb]{0.82,0.01,0.11}{\ \ \ \ \ \ 0.4+0.1}\\
\ \ \ \ \ +0.1
\end{array}$};

\end{tikzpicture}

%% file: supplementary.tex
\appendix
\setcounter{table}{0}
\renewcommand{\thetable}{A\arabic{table}}

\setcounter{figure}{0}
\renewcommand{\thefigure}{A\arabic{figure}}

\section*{Appendix}
In this appendix, we include the following additional details, which we could not include in the main paper due to space constraints.
\begin{itemize}
\setlength\itemsep{-0.1em}
    \item Proofs of propositions are presented in \S~\ref{proofsofpropositions}.
    \item Experimental setup is presented in \S~\ref{experimental setup}.
    \item Additional results and ablations studies are in \S~\ref{ablation}.
    \item Example of program synthesis application using FlashFill task is explained in \S~\ref{flashfill}.
    \item Example of a neurosymbolic program by solving XOR problem is discussed in \S~\ref{xor}
    \item Interpretability of Synthesized Programs using a real-world example is discussed in \S~\ref{appx_sec_interpretability}
    \item Additional related work based on discussed generative modeling-based causal effect estimation, causal discovery-based effect estimation, and the difference between neurosymbolic program synthesis and neural architecture search is discussed in  \S~\ref{sec additional related work}.
\end{itemize}

\section{Proofs of Propositions}
\label{proofsofpropositions}

\admissibility*

\begin{proof}
Let $\mathcal{G}$ denote the program graph that is being generated by an informed search algorithm. At any node $u$ in $\mathcal{G}$, let $s(u)$ be the structural cost of $u$ i.e., the sum of costs of rules used to construct $u$. Now, let $u[\alpha_1,\dots,\alpha_k]$ be any structure (that is not partial) obtained from $u$ by using the rules $\alpha_1,\dots,\alpha_k$. Then the cost to reach goal node from $u$ is given by:
\begin{equation*}
\begin{aligned}
J(u)=&\min_{\alpha_1,\dots,\alpha_n, \theta(u), \theta} [s(u(\alpha_1,\dots,\alpha_k))
- s(u) +\\
& \zeta(u[\alpha_1,\dots,\alpha_k], (\theta_u, \theta))]\\ 
\end{aligned} 
\end{equation*}
\noindent where $\theta(u)$ is the set of parameters of $u$ and $\theta$ is the set of parameters of $\alpha_1, \dots,\alpha_k$.
Now, let the heuristic function value $h(u)$ at $u$ be obtained as follows: substitute the non-terminals in $u$ with neural networks parametrized by the set of parameters $\omega$ (these networks are type-correct--- for example, if a non-terminal is supposed to generate sub-expressions whose inputs are sequences, then the neural network used in its place is recurrent). Now, let us denote the program obtained by this construction with $(\mathcal{P}(u), (\theta(u), \omega))$. The heuristic function value at $u$ is now given by:
\begin{equation*}
    h(u) = \min_{\theta(u), \omega} \zeta(\mathcal{P}(u), (\theta(u), \omega))
\end{equation*}
%\textbf{Side note:} 
\noindent In practice, neural networks may only form an approximate relaxation of the space of completions and parameters of architectures; also, the training of these networks may not reach global optima. To account for these issues, consider an approximate notion of admissibility~\cite{harris,heuristics}. For a fixed constant $\epsilon>0$, let an $\epsilon$-admissible heuristic be a function $h^{*}(u)$ over architectures such that $h^{*}(u) \leq J(u) + \epsilon; \forall u$. 

As neural networks with adequate capacity are universal function approximators, there exist parameters $\omega^{*}$ for our neurosymbolic program such that for all $u, \alpha_1,\dots,\alpha_k, \theta(u), \theta$:
\begin{equation*}
    \zeta(P(u), (\theta(u), \omega^*)) \leq \zeta(P(u[\alpha_1,\dots,\alpha_k]), (\theta(u), \theta)) + \epsilon
\end{equation*}
If $s(r)>0;\forall r\in \mathcal{L}$ (where $\mathcal{L}$ is the DSL under consideration), then $s(u) \leq s(u[\alpha_1,\dots,\alpha_k])$, which implies: 
\begin{equation*}
\begin{aligned}
h(u)&\leq \min_{\alpha_1,\dots,\alpha_n, \theta(u), \theta} \zeta(u[\alpha_1,\dots,\alpha_k], (\theta_u, \theta))) + \epsilon\\
    & \leq \min_{\alpha_1,\dots,\alpha_n, \theta(u), \theta} \zeta(u[\alpha_1,\dots,\alpha_k], (\theta_u, \theta)))\\
    & + s(u(\alpha_1,\dots,\alpha_k)) - s(u) + \epsilon\\
    &=  J(u) +\epsilon
\end{aligned}
\end{equation*}
In other words, $h(u)$ is $\epsilon$-admissible. 

Let $C$ denote the optimal path cost in $\mathcal{G}$. If an informed search algorithm returns a node $u_g$ as the goal node that does not have the optimal path cost $C$, then there must exist a node $u'$ on the frontier (nodes to explore) that lies along the optimal path but has not yet explored. Let $g(u_g)$ denote the path cost at $u_g$ (note that path cost includes the prediction error of the program at $u_g$). This lets us establish an upper bound on the path cost of $u_g$.
\begin{equation*}
    g(u_g) \leq g(u') + h(u') \leq g(u') + J(u') + \epsilon \leq C + \epsilon.
\end{equation*}
In an informed search algorithm, the heuristic estimate at the goal node $h(u_g)$ is 0. That is, the path cost of the optimal program returned by the informed search algorithm is at most an additive constant $\epsilon$ away from the path cost of the optimal solution.
\end{proof}

We apologize for the typo in the statement of Lemma 5.1 in the main paper. We hope the statement and the proof of Lemma 5.1 is clear in this Appendix. 

\guarantees*
\begin{proof} 

We know by universal approximation theorem~\cite{hornik} that there exist a trained 1-hidden layer neural network model $\mathcal{N}$ with $n$ inputs $x_1, \dots, x_n$, $d$ hidden neurons $h_1, \dots, h_d$, and output $\hat{y}$ that $\hat{\epsilon}$-approximates $g$ for some $\hat{\epsilon}>0$. We now show that $\mathcal{N}$'s output can be $\epsilon'$-approximated using a program synthesized using NPS with $\epsilon'$-admissible heuristic and a DSL.

In $\mathcal{N}$, let the activation function used in hidden and output layers be $f(\cdot)$; $\theta_{ij}$ be the weight connecting $i^{th}$ input to $j^{th}$ hidden neuron; and $\theta_j$ be the weight connecting $j^{th}$ hidden neuron to output $\hat{y}$. The output $\hat{y}$ of $\mathcal{N}$ can be expressed in terms of inputs, activation function, and parameters as:
\begin{equation}
\label{nn expression}
\begin{aligned}
    \hat{y}&=f(\theta_1 f(\theta_{11}x_1+\dots+\theta_{n1}x_n)+\dots\\
    &+\theta_d f(\theta_{1d}x_1 +\dots+\theta_{nd}x_n) )\\
\end{aligned}
\end{equation}
Since the above expression consists of additions, multiplications, and an activation function $f$, it is easy to see that Eqn~\ref{nn expression} can be synthesized using the following DSL $\mathcal{L}$:
\begin{equation*}
    \mathtt{\alpha \coloneqq f(\alpha)\ |\ mul(\theta, \alpha)\ |\ add(\alpha,\alpha)\ |\ x_1\ |\ \dots\ |\ x_n }
\end{equation*}
where $\mathtt{mul, add}$ represent usual multiplication and addition operations. If $d=2$ and $n=2$, the synthesized program that matches the expression for $y$ in Eqn~\ref{nn expression} looks like: 
\begin{equation}
\label{npsexpression}
\begin{aligned}
& \mathtt{f\ (add\ (mul\ (\theta,\  f(\ add\ (mul\ (\theta,\ x_1),mul(\theta,\ x_2)))),} \ \\
&\mathtt{mul(\theta,\ f\ (add\ (mul(\theta,\ x_1), mul(\theta,\ x_2))))))}
\end{aligned}
\end{equation}
Note that $\theta$ is overloaded in the above expression only for convenience and readability; each $\theta$ is however updated independently while training the above program using gradient descent. 

Using Expression~\ref{npsexpression}, it is clear that  Eqn~\ref{nn expression} can be synthesized using $\mathcal{L}$ for any given $n, d$. Now, as part of our construction, set $s(r)=0; \forall r\in \mathcal{L}$ to synthesize programs of arbitrary depth and width without worrying about the structural cost of the synthesized program. Now the path cost $p$ of a node $u$ returned by the synthesizer contains only the prediction error value of the program at the node $u$ (Eqn ~\ref{objective}). Using Lemma~\ref{prop1}, $p$ is at most $\epsilon'$ away from the path cost of the optimal solution (node with the expression for $y$, the output of $\mathcal{N}$). Since the path cost of any node only contains the prediction error values, we conclude that the loss incurred by the synthesized program is $\epsilon'-$close to the loss incurred by $\mathcal{N}$.

Finally, as per the universal approximation theorem, we can increase the number of hidden layer neurons of a 1-hidden layer NN $\mathcal{N}$ to approximate $g$ with a certain error, say $\hat{\epsilon}$. Also, there exists a neurosymbolic program $(\mathcal{P},\theta)$ whose error in approximating $\mathcal{N}$ is $\epsilon'$. Equivalently, there exists a neurosymbolic program $(\mathcal{P},\theta)$ whose error in approximating $f$ is $(\hat{\epsilon}+\epsilon')$. If we choose $\hat{\epsilon}, \epsilon'$ such that $\epsilon = \hat{\epsilon}+\epsilon'$ for a given $\epsilon$, we have the desired result.
\end{proof}

Before proceeding with the proof of next proposition, we describe how NESTER, using the proposed DSL can generate programs $(\mathcal{P}_{C},\theta_{C})$, and $(\mathcal{P}_{D},\theta_{D})$, whose architectures are similar to that as CFR~\cite{tarnet} and Dragonnet~\cite{dragon_net}. The program architectures for $\mathcal{P}_C,$ $\mathcal{P}_D$ are shown in Fig~\ref{fig:cfrcomparison}, ~\ref{fig:dragonnetcomparison}. The parameter sets $\theta_C,$ and $\theta_D$ are implicit in the program primitives used in the respective architectures. 

\noindent \textbf{Construction of $\mathcal{P}_C$:}
CFR is an extension of TARNet as explained below. TARNet is a simple 2-head network without any constraints on the learned representation $\phi$ (Fig~\ref{fig:comparison}). CFR minimizes the distance between $\phi(\mathbf{x}|t=1), \phi(\mathbf{x}|t=1)$ to achieve IPM regularization. To get similar behavior, $\mathcal{P}_C$ uses $\mathtt{align}$ primitive that implicitly generates representations close to each other for inputs with different treatment values. $\mathcal{P}_C$ has two heads corresponding to two $\mathtt{mlp}$ primitives that output treatment-specific effects. 

\noindent \textbf{Construction of $\mathcal{P}_D$:}
In Dragonnnet~\cite{dragon_net}, along with two treatment-specific heads (similar to TARNet), another head predicts the treatment variable so that the parents of the treatment variable are being used for propensity score matching. To achieve this behavior, $\mathcal{P}_D$ uses $\mathtt{propensity}$ primitive that controls propensity score, which is the motivation behind predicting treatment variable from the representation $\phi$. $\mathcal{P}_D$ has two heads similar to Dragonnet to predict the treatment-specific effects.
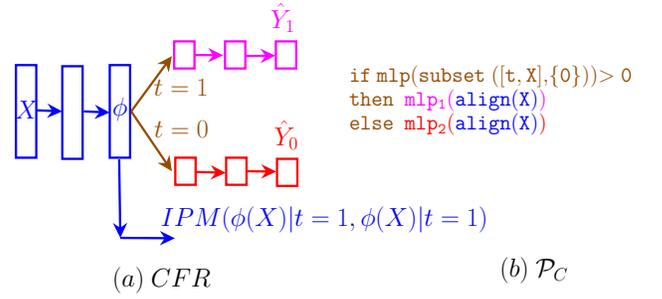
\begin{figure}%[28]{r}{0.55\textwidth}
    \centering
    \scalebox{0.57}{
    \tikzset{every picture/.style={line width=0.75pt}} %set default line width to 0.75pt  
    \input{images/cfr.tikz}
}
    \caption{(a) CFR architecture (b) Program $\mathcal{P}_C$ synthesized by NESTER using our DSL (Tab~\ref{tab:dsl_causal_inference}) that is functionally similar to CFR. Colors are used to show the equivalence between the components of CFR and $\mathcal{P}_C$.}
\label{fig:cfrcomparison}
\end{figure}
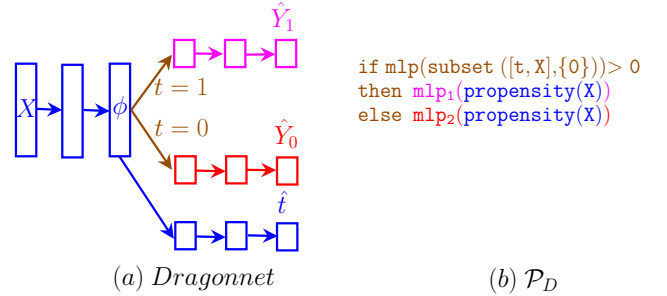
\begin{figure}%[28]{r}{0.55\textwidth}
    \centering
    \scalebox{0.57}{
    \tikzset{every picture/.style={line width=0.75pt}} %set default line width to 0.75pt  
    \input{images/dragonnet.tikz}
}
    \caption{(a) Dragonnet architecture (b) Program $\mathcal{P}_D$ synthesized by NESTER using our DSL (Tab~\ref{tab:dsl_causal_inference}) that is functionally similar to Dragonnet. Colors are used to show the equivalence between the components of Draonnet and $\mathcal{P}_D$.}
\label{fig:dragonnetcomparison}
\end{figure}

\reduction*
\begin{proof}

Since CFR provides error bounds in estimating $\epsilon_{PEHE}$, we show how such bounds can be extended to NESTER. We first restate the following definitions and notations from~\cite{tarnet}. 

Let $p^{t=1}(\mathbf{x})= p(\mathbf{x}|t=1)$, and $p^{t=0}(\mathbf{x})= p(\mathbf{x}|t=0)$ denote respectively the treatment and control distributions. Let $\phi: \mathbf{X} \rightarrow \mathcal{R} $ be the representation function which is assumed to be one-to-one and differentiable. Let $p^{t=1}_{\phi}(\mathbf{x})= p_{\phi}(\mathbf{x}|t=1)$, and $p^{t=0}_{\phi}(\mathbf{x})= p_{\phi}(\mathbf{x}|t=0)$ denote respectively the treatment and control distributions induced over $\mathcal{R}$. Let $h:\mathcal{R}\times \{0,1\} \rightarrow Y$ be a hypothesis function (e.g., treatment-specific heads of TARNet/CFR). The expected loss for the unit $(\mathbf{x}, t)$ is defined as follows
\begin{equation*}
    l_{h,\phi}(\mathbf{x}, t) \coloneqq \int_{Y}L(Y^t, h(\phi(\mathbf{x}),t))p(Y^t|\mathbf{x}) dY^t
\end{equation*}
Where $L : Y \times Y \rightarrow \mathcal{R}^+$ is squared loss function defined as $L(y,\hat{y})\coloneqq (y-\hat{y})^2$. Now consider the two complimentary loss functions: one is the standard machine learning loss, call the factual loss, denoted by $\epsilon_F$. The other is the expected loss
with respect to the distribution where the treatment assignment is flipped, called the counterfactual loss, $\epsilon_{CF}$. These are defined as follows

\begin{equation*}
        \epsilon_F(h,\phi) \coloneqq  \int_{\mathbf{X}\times\{0,1\}} l_{h,\phi}(\mathbf{x},t)p(\mathbf{x},t)d\mathbf{x}dt
\end{equation*}
\begin{equation*}
        \epsilon_{CF}(h,\phi) \coloneqq \int_{\mathbf{X}\times\{0,1\}} l_{h,\phi}(\mathbf{x},t)p(\mathbf{x},1-t)d\mathbf{x}dt
\end{equation*}

Similarly, one can define the expected treated and control losses as follows
    \begin{equation*}
        \epsilon_F^{t=1} (h,\phi) = \int_{\mathbf{X}} l_{h,\phi}(\mathbf{x},1)p^{t=1}(\mathbf{x}) d\mathbf{x}
\end{equation*}
\begin{equation*}
        \epsilon_F^{t=0} (h,\phi) = \int_{\mathbf{X}} l_{h,\phi}(\mathbf{x},0)p^{t=0}(\mathbf{x}) d\mathbf{x}
\end{equation*}

\begin{equation*}
       \epsilon_{CF}^{t=1} (h,\phi) = \int_{\mathbf{X}} l_{h,\phi}(\mathbf{x},1)p^{t=0}(\mathbf{x}) d\mathbf{x}
\end{equation*}
\begin{equation*}
        \epsilon_{CF}^{t=0} (h,\phi) = \int_{\mathbf{X}} l_{h,\phi}(\mathbf{x},0)p^{t=1}(\mathbf{x}) d\mathbf{x}
\end{equation*}

For $u \coloneqq p(t = 1)$, we have the following~\cite{tarnet}
\begin{equation*}
    \begin{aligned}
        \epsilon_F(h,\phi) &= u\epsilon_F^{t=1}(h,\phi) + (1-u)\epsilon_F^{t=0}(h,\phi)\\
        \epsilon_{CF}(h,\phi) &= (1-u)\epsilon_{CF}^{t=1}(h,\phi) + u\epsilon_{CF}^{t=0}(h,\phi)\\
    \end{aligned}
\end{equation*}
Let $G$ be a function family consisting of functions $g:\mathcal{S}\rightarrow \mathbb{R}$. For a pair of distributions $p_1, p_2$ over $\mathcal{S}$, the Integral Probability Metric is defined as follows
\begin{equation*}
    IPM_G(p_1, p_2) =  \underset{g \in G}{sup} \bigl\lvert \int_{\mathcal{S}} g(s) (p_1(s)-p_2(s)) ds \bigr\rvert
\end{equation*}
For $t\in \{0,1\}$, let $m_t(\mathbf{x}) = E [Y^t|\mathbf{x}]$, $\tau(\mathbf{x}) = m_1(\mathbf{x}) - m_0(\mathbf{x})$ and $\hat{\tau}(\mathbf{x}) = f(\mathbf{x},1) - f(\mathbf{x},0)$ ($f$ is defined in Sec.~\ref{sec_background}). Then we have the following
\begin{equation*}
    \epsilon_{PEHE}(f) \coloneqq \int_{\mathbf{X}} (\hat{\tau}(\mathbf{x})-\tau(\mathbf{x}))^2p(\mathbf{x})d\mathbf{x}
\end{equation*}
Let 
$$\sigma^2_{Y^t} (p(\mathbf{x}, t)) = \int_{\mathbf{X}\times Y} (Y^t - m_t(\mathbf{x}))^2 p(Y^t|\mathbf{x})p(v, t) dY^t d\mathbf{x}$$ 

and $\sigma^2_{Y^t} = min\{\sigma^2_{Y^t}(p(\mathbf{x}, t)), \sigma^2_{Y^t}(p(\mathbf{x}, 1-t))\}$ 

and $\sigma^2_Y = min \{\sigma^2_{Y^0}, \sigma^2_{Y^1}\}$

Now assume there exists a constant $B_{\phi}$ and loss $L(y_1,y_2) = (y_1-y_2)^2$ such that for $t\in\{0,1\}$, the functions $g_{\phi, h}(r, t)\coloneqq \frac{1}{B_{\phi}}l_{h,\phi}(\psi(r), t)\in G$. Then we have
\begin{equation}
\label{bounds}
\begin{aligned}
      &\epsilon_{PEHE} (h,\phi) \leq 2(\epsilon_{CF} (h,\phi)+\epsilon_F (h,\phi)-2\sigma^2_Y) \\
  &\leq 2(\epsilon_F^{t=0}(h,\phi) + \epsilon_F^{t=1}(h,\phi)+ B_{\phi} IPM_G(p_{\phi}^{t=0}, p_{\phi}^{t=1})-2\sigma^2_Y) 
\end{aligned}
\end{equation}
We refer to ~\cite{tarnet} for the complete proof of the inequality~\ref{bounds}, which is valid for the Counterfactual Regression (CFR) model~\cite{tarnet}. We now present the following equivalences to show that the above error bound is valid for the program $(\mathcal{P}_C, \theta_C)$ equivalent to CFR.

\begin{itemize}
    \item In $\mathcal{P}_C$, $\mathtt{subset(\mathbf{v}, \{0\})}$ acts as the decision node to decide which specific $\mathtt{mlp(align(\mathtt{v}))}$ to execute.  These specific $\mathtt{mlp(align(\mathtt{v}))}$ are the same as the outputs of the hypothesis function $h$ used in the factual and counterfactual losses $\epsilon_F, \epsilon_{CF}$ defined earlier.
    \item By our construction of $(\mathcal{P}_C, \theta_C)$, we have two $\mathtt{mlp(align(\mathtt{v}))}$ primitives to output $p^{t=0}_{\phi}$ and $p^{t=1}_{\phi}$. $\phi$ is trained to minimize the MMD between $p^{t=0}_{\phi}$ and $p^{t=1}_{\phi}$. Since MMD is one specific IPM, we replace IPM with MMD in the inequality~\ref{bounds}.
    \item $\sigma^2_Y$ can be directly obtained from the observational data. Hence the error bounds guaranteed by NESTER w.r.t. $\epsilon_{PEHE}$ is as follows.
    \begin{equation*}
    \label{bounds_nester}
    \small{
    \begin{aligned}
        &\epsilon_{PEHE} (h,\phi)
        \leq 2(\epsilon_{CF} (h,\phi)+\epsilon_F (h,\phi)-2\sigma^2_Y)\\
        &\leq 2(\epsilon_F^{t=0}(h,\phi) + \epsilon_F^{t=1}(h,\phi)
        + B_{\phi} MMD(p_{\phi}^{t=0}, p_{\phi}^{t=1})-2\sigma^2_Y)
    \end{aligned}
    }
    \end{equation*}
    
\end{itemize}
\end{proof}

\section{Algorithm}
\label{sec algorithm}
Algorithm~\ref{nester} outlines the procedure for obtaining a program structure to estimate causal effects.

\begin{algorithm}
    \caption{NESTER using $A^*$}
    \label{nester}
\begin{algorithmic}[1]
  \Require Root node $u_0$ with initial non-terminal $\rho$, DSL $\mathcal{L}$.

\State {\bfseries Initialize:}
     $Q=\{u_0\}$, $f(u_0)= \infty$, $u_{best}=u_0$, $f=\infty$

\While{$Q \neq \emptyset$}
    
    \State $v = \argmin_{u\in Q} f(u)$
    \State $Q = Q\setminus \{v\}$\;
    
   \If{$v$ is leaf node and $f(v)<f$}
    \State $f = f(v)$
    \State $u_{best} = v$
  \Else
    \For{each child $u$ of $v$}
    
    \State $h(u) = \min_{\theta(u)} \zeta(\mathcal{P}(u), \theta(u))$ 

    \State $f(u) = s(\mathcal{P}(u))+h(u)$
    
    \State $Q = Q\cup \{u\}$
    \EndFor
\EndIf
\EndWhile
\State return $u_{best}$ 
\end{algorithmic}
\end{algorithm}

\section{Experimental Setup}
\label{experimental setup}
\subsection{Details on Datasets}

\noindent \textbf{IHDP:} Infant Health and Development Program (IHDP) is a randomized control experiment on 747 low-birth-weight, premature infants. The treatment group consists of 139 children, and the control group has 608 children. The treatment group received additional care such as frequent specialist visits, systematic educational programs, and pediatric follow-up. The Control group only received pediatric follow-up.~\cite{ihdp_dataset} created the semi-synthetic version of IHDP dataset by synthesizing both potential outcomes. Following ~\cite{ihdp_dataset, tarnet, yoon2018ganite, dragon_net}, we use simulated outcomes of the IHDP dataset from NPCI package~\cite{npci}. This experiment aims to estimate the effect of treatment on the IQ score of children at the age of 3.
\begin{table*}
\centering
\scalebox{0.95}{
    \begin{tabular}{cccccc}
    \toprule
         Dataset & Sample Size & Input Size & Batch Size & Epochs &Train/Valid/Test Split (\%)\\
        \midrule
        IHDP&747 (1000 such instances)&26&16&100&64/16/20\\
        Twins&11400&31&128&7&64/16/20\\
        Jobs&3212&18&64&10&64/16/20\\
         \bottomrule
    \end{tabular}
    }
    \caption{Dataset details. `Input Size' includes the treatment variable.}
        \label{tab:experimental_setup}
\end{table*}

\noindent \textbf{Twins:} 
The Twins dataset is derived from all births in the USA between 1989-1991~\cite{twins}. Considering twin births in this period, for each child, we estimate the effect of birth weight on 1-year mortality rate. Treatment $t = 1$ refers to the heavier twin and $t = 0$ refers to the lighter twin. Following~\cite{yoon2018ganite}, for each twin-pair, we consider 30 features relating to the parents, the pregnancy, and the birth. We only consider twins weighing less than 2kg and without missing features. The final dataset has 11,400 pairs of twins whose mortality rate for the lighter twin is $17.7\%$, and for the heavier $16.1\%$. In this setting, for each twin pair we observed both the case $t = 0$ (lighter twin) and $t = 1$ (heavier twin) (that is, since all other features such as parent's race, health status, gestation weeks prior to birth, etc. are same except the weight of each twin, the choice of twin (lighter vs. heavier) is associated with the treatment ($t=0\ vs\ t=1$)); thus, the ground truth of individualized causal effect is known in this dataset. To simulate an observational study from these 11,400 pairs, following~\cite{yoon2018ganite}, we selectively observe one of the two twins using the feature information $\mathbf{x}$ (to create selection bias) as follows: $t|\mathbf{x} \sim$ $\texttt{Bernoulli}$($\texttt{sigmoid}$($\mathbf{w}^T \mathbf{x} + n$)) where $\mathbf{w}^T \sim U((-0.1, 0.1)^{30\times 1})$ and $n \sim N (0, 0.1)$.

\noindent \textbf{Jobs:}
The Jobs dataset is a widely used real-world benchmark dataset in causal inference. In this dataset, the treatment is job training, and the outcomes are income and employment status after job training. The dataset combines a randomized study based on the National Supported Work Program in the USA (we denote the observations from this randomized study with $E$) with observational data~\cite{jobs1}. Each observation contains 18 features: age, education, previous earnings, etc. Following~\cite{tarnet, yoon2018ganite}, we construct a binary classification task, where the goal is to predict unemployment status given a set of features. The Jobs dataset is the union of 722 randomized samples ($t = 1: 297, t = 0: 425$) and 2490 observed samples $(t = 1: 0, t = 0: 2490)$. The treatment variable is job training ($t=1$ if trained for job else $t=0$), and the outcomes are income and employment status after job training. In Eqns~\ref{eqn_att}-\ref{eqn_att_true}, we then have $|T|=297, |C| = 2915, |E|=722$. Since all the treated subjects $T$ were part of the original randomized sample E, we can compute the true $ACT$ (Eqn~\ref{eqn_att_true}) and hence can study the precision in the estimation of $ACT$ (Eqn~\ref{eqn_att}). Tab~\ref{tab:experimental_setup} summarizes the dataset details.

\subsection{Additional Details on Evaluation Metrics}
For the experiments on IHDP and Twins datasets where we have access to both potential outcomes, following~\cite{tarnet,yoon2018ganite,dragon_net,balance_reg}, we use the evaluation metrics: \textit{error in the estimation of Average Causal Effect ($\epsilon_{ACE}$)} and the \textit{expected Precision in Estimation of Heterogeneous Effect ($\epsilon_{PEHE}$)}. For a sample of $n$ data points, $\epsilon_{ACE}$, $\epsilon_{PEHE}$ are defined as follows. 
\begin{align*}
    \epsilon_{ACE} &\coloneqq |\frac{1}{n}\sum_{i=1}^n [f(\mathbf{x}_i,1)-f(\mathbf{x}_i,0)] - \frac{1}{n}\sum_{i=1}^n[Y^1_i-Y^0_i]| \\
    \epsilon_{PEHE} &\coloneqq \frac{1}{n}\sum_{i=1}^n [(f(\mathbf{x}_i,1)-f(\mathbf{x}_i,0)) - (Y^1_i-Y^0_i)]^2 
\end{align*}
For the experiment on the Jobs dataset where we observe only one potential outcome per data point, following~\cite{tarnet,yoon2018ganite,dragon_net,balance_reg}, we use the metric \textit{error in estimation of Average Causal Effect on the Treated} ($\epsilon_{ACT}$), which is defined as follows.
\begin{align}
    \label{eqn_att}
    \epsilon_{ACT} &\coloneqq |ACT^{true} - \frac{1}{|T|} \sum_{i\in T}[f(\mathbf{x}_i,1)-f(\mathbf{x}_i,0)]|
\end{align}
where $ACT^{true}$ is defined as:
\begin{align}
\label{eqn_att_true}
    ACT^{true} &\coloneqq \frac{1}{|T|} \sum_{i \in T} Y^1_i - \frac{1}{|U\cap E|} \sum_{i\in U\cap E} Y^0_i
\end{align}
and $T$ is the treated group, $U$ is the control group, and $E$ is the set of data points from a randomized experiment~\cite{tarnet} (see  description of Jobs dataset below for an example of $E,T,$ and $U$).

\noindent \textbf{Understanding $k$-NN results:} In $k$-NN algorithm, if $k=1$ and treatment value $t=1$, $f(\mathbf{x}_i,1)$ is exactly same as $Y^1_i$. If treatment value $t=0$, $f(\mathbf{x}_i,0)$ is exactly same as $Y^0_i$ because of the way k-NN works during test time on in-sample data. For this reason, the estimated value of $\epsilon_{ACE}$ is biased towards $0$. This bias exists even for higher values of $k$ in $k$-NN while taking the average outputs of $k$ nearest data points. However, we do not observe such bias w.r.t. out-sample data. Hence, following earlier work~\cite{yoon2018ganite}, we only consider k-NN results for out-sample performance.

\section{Additional Empirical Analysis and Discussion}
\label{ablation}
 
This section presents ablation studies to understand various aspects of synthesized programs.

\noindent \textbf{Interpretability of NESTER:}
We now provide the interpretability to the program generated by NESTER for estimating causal effects for the Twins dataset: ``$\mathtt{subset(\mathbf{v},\{0..|\mathbf{v}|\})}$'', which is same as $\mathtt{mlp(\mathbf{v})}$ because all the features of $\mathtt{\mathbf{v}}$ are chosen by $\mathtt{subset}$ primitive. Since the $\mathtt{subset}$ primitive allows us to check the performance w.r.t. different subsets of covariates, we empirically verified the effect of choosing a subset of input covariates (other covariates are set to $0$) on $\epsilon_{ACE}$. 

\begin{figure}[H]
    \centering
    \includegraphics[width=0.30\textwidth]{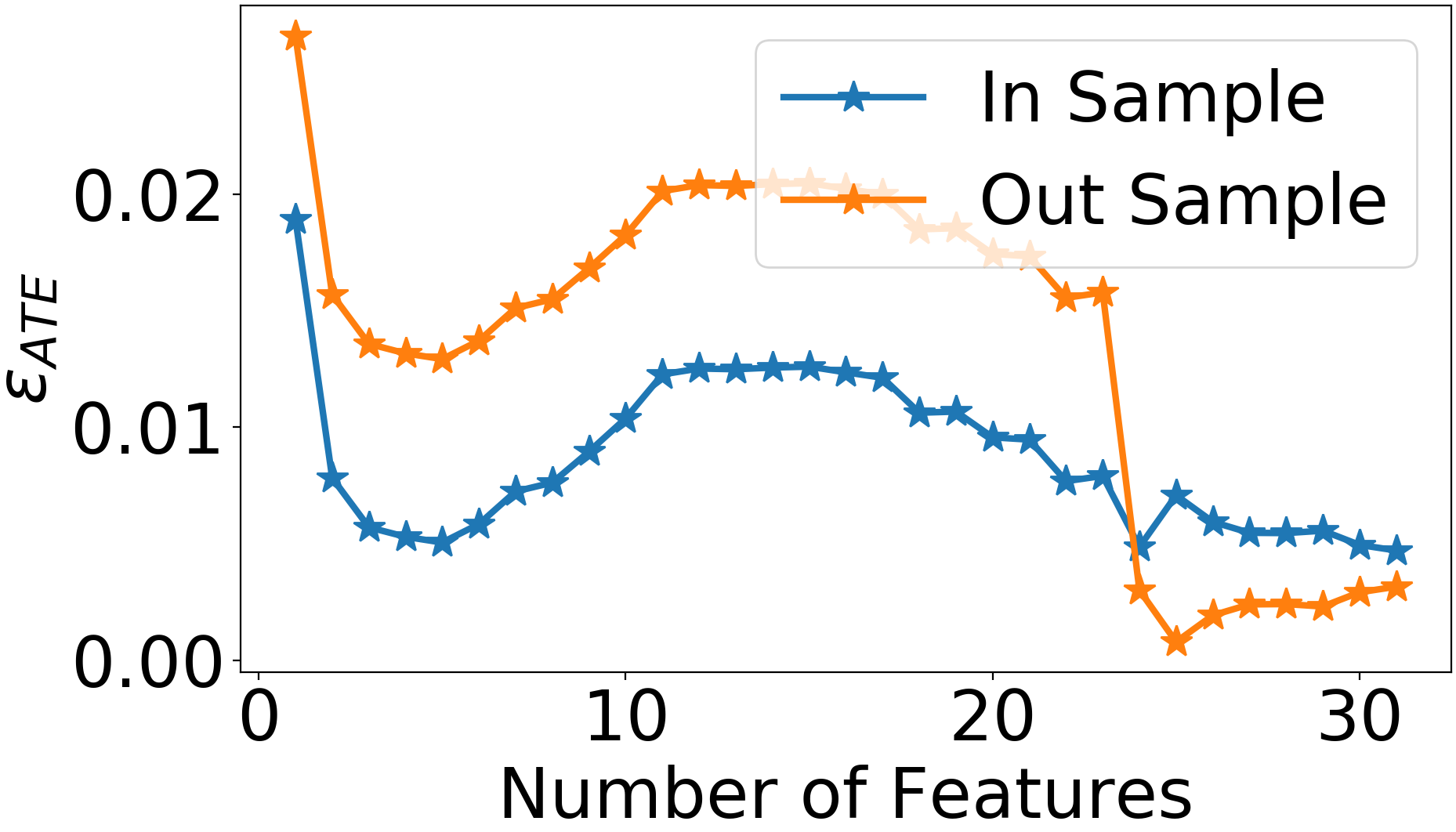}
    \caption{ Feature count vs $\epsilon_{ACE}$}
    \label{twins_analysis}
\end{figure}

Results in Fig~\ref{twins_analysis} show the performance of NESTER as the number of covariates are increased from 1 to 31 (starting with treatment variable, adding one covariate at a time; we chose this ordering because checking with all possible subsets, $2^{31}$ in this case, is not feasible). This analysis allows us to understand which features are useful to better estimate causal effects and which are not. For example, if the exclusion of a set of features $S$ is improving a model's performance in estimating causal effects, $S$ can then be used to infer the causal relationships in the underlying causal structure of the data (e.g., if the sample size is small, not controlling the parents of treatment variable that are not the ancestors of the target variable is likely to improve the precision in the estimation of causal effects~\cite{crashcourse}). Also, this simple program synthesized by NESTER supports the fact that simpler models perform better on the Twins dataset. This can be observed from the first three rows and the final row of Tab~\ref{tab:results}.

\begin{table}
    \centering
    \label{tab:results_pehe}
    \scalebox{0.75}{
\begin{tabular}{lcccc}
    \toprule
    Datasets (Metric) $\mathbf{\rightarrow}$&\multicolumn{2}{c}{IHDP ($\sqrt{\epsilon_{PEHE}} (\downarrow)$)}&\multicolumn{2}{c}{Twins ($\sqrt{\epsilon_{PEHE}} (\downarrow)$)} \\
    \midrule
       Methods $\mathbf{\downarrow}$ &In-Sample&Out-Sample&In-Sample&Out-Sample\\
    \midrule
        OLS-1 &5.80$\pm$0.30&5.80$\pm$0.30&.319$\pm$.001&.318$\pm$.007\\
        
        OLS-2 &2.50$\pm$0.10&2.50$\pm$0.10&.320$\pm$.002&.320$\pm$.003\\
        
        k-NN &2.10$\pm$0.10&4.10$\pm$0.20&.333$\pm$.001&.345$\pm$.007\\

        BLR&5.80$\pm$0.30&5.80$\pm$0.30&.312$\pm$.003&.323$\pm$.018\\
    \midrule
        BART &2.10$\pm$0.10&2.30$\pm$0.10&.347$\pm$.009&.338$\pm$.016\\
        
        R Forest &4.20$\pm$0.20&6.60$\pm$0.30&.306$\pm$.002&.321$\pm$.005\\
        
        C Forest &3.80$\pm$0.20&3.80$\pm$0.20&.366$\pm$.003&.316$\pm$.011\\
    \midrule
        BNN &2.20$\pm$0.10&2.10$\pm$0.10&.325$\pm$.003&.321$\pm$.018\\
        
        TARNet &0.88$\pm$0.02&0.95$\pm$0.02&.317$\pm$.005&.315$\pm$.003\\
        
        MHNET&1.54$\pm$0.70&1.89$\pm$0.52&.319$\pm$.000&.321$\pm$.000\\
        
        GANITE&1.90$\pm$0.40&2.40$\pm$0.40&\textbf{.289$\pm$.005}&\textbf{.297$\pm$.016}\\
        
        CFR$_{WASS}$ & 0.71$\pm$0.02&\textbf{0.76$\pm$0.02}&.315$\pm$.007&.313$\pm$.003\\
        
        Dragonnet &1.37$\pm$1.57&1.42$\pm$1.67&.319$\pm$.000&.321$\pm$.000\\
        
        CMGP &\textbf{0.65$\pm$0.44}&0.77$\pm$0.11&.320$\pm$.002&.319$\pm$.008\\
        TNet&0.90$\pm$0.01&0.91$\pm$0.03&.318$\pm$.002&.319$\pm$.000\\
        SNet&0.69$\pm$0.01&\textbf{0.76$\pm$0.01}&.318$\pm$.002&.318$\pm$.000\\
    \midrule
        \textbf{NESTER - NEAR} &0.73$\pm$0.19&\textbf{0.76$\pm$0.20}&.318$\pm$.002&.319$\pm$.000\\
        \textbf{NESTER - dPads}&0.71$\pm$0.10&\textbf{0.76$\pm$0.32}&.314$\pm$.001&.331$\pm$.001\\
        \bottomrule
    \end{tabular}
    }
    \caption{ Results on IHDP, Twins dataset on $\epsilon_{PEHE}$}
    
\end{table}

\begin{figure}
    \centering
    \includegraphics[width=0.48\textwidth]
    {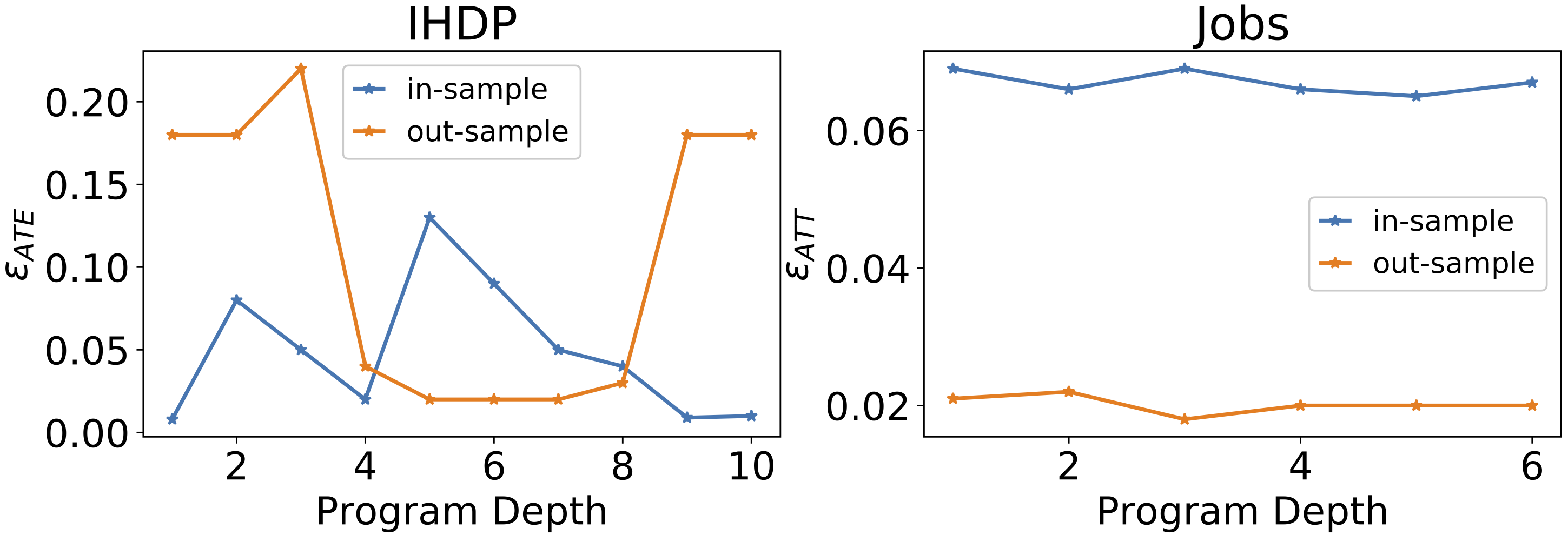}
    \caption{ Program depth vs performance.}
    \label{program_depth}
\end{figure}

\noindent \textbf{Analysis on Depth of Synthesized Program Structures:}
We study the effect of program depth on the estimated causal effects while keeping all other hyperparameters fixed. Fig~\ref{program_depth} shows the results on IHDP and Jobs datasets for various values of program depth. Since IHDP dataset contains 1000 realizations of simulated outcomes~\citep{ihdp_dataset}, we take the first realization and verify the effect of program depth on $\epsilon_{ACE}$. For a program depth of 4, we observed that the sum of in-sample and out-sample $\epsilon_{ACE}$ is less compared to other depths. We believe that this is because of model over-fitting for large program depths (In Fig~\ref{program_depth} left, out-sample $\epsilon_{ACE}$ is increasing while in-sample $\epsilon_{ACE}$ is decreasing). In the Jobs dataset, we observed that almost all program depths results in similar in-sample and out-sample $\epsilon_{ACT}$. Hence, in this case it is advisable to limit the program depth to be a small number as it helps to interpret the results better. On Twins dataset, we observed that simple models give best results. It is observed that, even though we set the hyperparameter that controls the depth of the program graph to be a large value, the resultant optimal program always ends up to be of depth 1, again supporting our claim that simple models work better for the Twins dataset.

\begin{table}
    \centering
    \scalebox{0.76}{
    \begin{tabular}{lccc}
    \toprule
    Dataset&SNet&NESTER-NEAR&NESTER-dPads \\
    \midrule
        Twins &1.85$\pm$0.3&0.30$\pm$0.01&\textbf{1.40$\pm$0.20}\\
        Jobs &1.23$\pm$0.2&1.09$\pm$0.40&\textbf{1.00$\pm$0.10}\\
    \bottomrule
    \end{tabular}
    }
        \caption{Run time in minutes}
    \label{tab:runtime}

\end{table}
\noindent \textbf{Runtime Analysis:} We also compare the run time of NESTER against the state-of-the-art learning-based method SNet~\citep{curth2021nonparametric}. As shown in Tab~\ref{tab:runtime}, on Twins and Jobs datasets, NESTER-dPads require less time than SNet and NESTER-NEAR on average. NESTER-NEAR also achieves state-of-the-art performance with smaller program depths, avoiding heavy computation requirements in practice. Experiments are conducted on a computing unit with an NVIDIA Tesla V100, and the average time over ten runs is reported.
\section{FlashFill Task and Semantics of its DSL}
\label{flashfill}

Following our discussion in \S~\ref{sec introduction} of the main paper, for better understanding of symbolic program synthesis, we provide an example of a symbolic program application called FlashFill~\cite{parisotto2016neuro}. Examples of the FlashFill task and a DSL to synthesize programs that solve FlashFill task are given in Tab~\ref{tab:flashfill}. Semantics of the DSL in Tab~\ref{tab:flashfill} Right are as follows.
\begin{table}
	\begin{minipage}{0.5\linewidth}
 \centering
         \begin{tabular}{ll}
            \toprule
            Input&Output\\
            \midrule
             William Henry Charles & Charles, W. \\
             Michael Johnson&Johnson, M.\\
             Barack Rogers&Rogers, B.\\
             Martha D. Saunders& Saunders, M.\\
            Peter T Gates     & Gates, P.\\ 
            \bottomrule
            \end{tabular}
	    \label{tab:flashfill1}
	\end{minipage}
	\begin{minipage}{0.5\linewidth}
 \centering
    \begin{align*}
     &\mathtt{String\ e:=}\mathtt{\ Concat(f_1,\dots,f_n)}\\
     &\mathtt{Substring\ f:=}\mathtt{\ ConstStr(s)|SubStr(v,p_l,p_r)}\\
     &\mathtt{Position\ p:=}\mathtt{\ (r, k, dir)|ConstPos(k)}\\
     &\mathtt{Direction\ Dir\ :=}\mathtt{\ Start|End}\\
     &\mathtt{Regex\ r\ :=}\mathtt{\ s|T_1|\dots|T_n}\\
    \end{align*}
	\end{minipage}
 \caption{\textbf{Top}: An example FlashFill task where input names are automatically translated to an output format in which last name is followed by the initial of the first name; \textbf{Bottom}: The DSL for FlashFill task based on regular expression string transformations~\cite{parisotto2016neuro}.}
	\label{tab:flashfill}

\end{table}
\begin{itemize}

\setlength\itemsep{-0.1em}
    \item $\mathtt{Concat(f_1,\dots,f_n)}$ - concatenates the results of the expressions $\mathtt{f_1,\dots,f_n}$.
    \item $\mathtt{ConstStr(s)}$ - returns the constant string $\mathtt{s}$.
    \item $\mathtt{SubStr(v,p_l,p_r)}$ - returns substring $\mathtt{v[p_l..p_r]}$ of the string $\mathtt{v}$, using position logic corresponding to $\mathtt{p_l,p_r}$. $\mathtt{v[i..j]}$ denotes the substring of string $\mathtt{v}$ starting at index $\mathtt{i}$ (inclusive) and ending at index $\mathtt{j}$ (exclusive), and $\mathtt{len(v)}$ denotes the length of the string $\mathtt{v}$
    \item $\mathtt{ConstPos(k)}$ - returns $\mathtt{k}$ if $\mathtt{k\geq 0}$ else return $\mathtt{l+k}$ where $\mathtt{l}$ is the length of the string
    \item $\mathtt{(r, k, Start)}$ -  returns the Start of $\mathtt{k^{th}}$ match of the expression $\mathtt{r}$ in $\mathtt{v}$ from the beginning (if $\mathtt{k \geq  0}$) or from the end (if $\mathtt{k < 0}$).
    \item $\mathtt{(r, k, End)}$ -  returns the End of $\mathtt{k^{th}}$ match of the expression $\mathtt{r}$ in $\mathtt{v}$ from the beginning (if $\mathtt{k \geq 0}$) or from the end (if $\mathtt{k < 0}$).
\end{itemize}
Based on the above semantics, a program that generates the desired output given the input names in Tab~\ref{tab:flashfill} is: $\mathtt{Concat(f_1, ConstStr(","), f_2, ConstStr("."))}$ where $\mathtt{f_1 \equiv SubStr(v, ("\ ", -1, End), ConstPos(-1))}$ and $\mathtt{f_2 \equiv SubStr(v, ConstPos(0), ConstPos(1))}$.

\section{Neurosymbolic Program Example: Solving XOR Problem} 
\label{xor}

Following our discussion in \S~\ref{sec_background} of the main paper, for better understanding of the internal workings of a neurosymbolic program, we provide an example on solving the XOR problem i.e., predicting the output of XOR operation given two binary digits. Unlike symbolic programs, neurosymbolic programs are differentiable and can be trained using gradient descent. Program primitives in a neurosymbolic program have trainable parameters associated with them. The program shown in Eq~\ref{eq xor} is constructed using (i) $\mathtt{if-then-else}$ and (ii) $\mathtt{affine}$ program primitives. 

\begin{equation}
\label{eq xor}
\begin{aligned}
            &\mathtt{if \ affine_{[1,1;0]}(x) >0 \ then }\\
            &\hspace{20pt} \mathtt{if\ affine_{[1,1;-1]}(x)>0 \ then }\\
            &\hspace{40pt} \mathtt{affine_{[0,0;0]}(x)}\\
            &\hspace{20pt} \mathtt{else}\\
            &\hspace{40pt} \mathtt{affine_{[1,1;0]}(x) }\\
            & \mathtt{else}\\
            &\hspace{20pt} \mathtt{ affine_{[0,0;0]}(x)}
\end{aligned}    
\end{equation}

$\mathtt{affine}$ primitive takes a vector as input and returns a scalar that is the sum of dot product of parameters with the input and a bias parameter. For example, if $\mathtt{x}=[1,0]$ then $\mathtt{affine_{[\theta_1,\theta_2;\theta_3]}(x) = \theta_1\times1+\theta_2\times0+\theta_3} = \theta_1+\theta_3$. The subscripts of $\mathtt{affine}$ in $\mathtt{affine_{[\theta_1,\theta_2;\theta_3]}}$ contain the parameters $\theta_1, \theta_2$ and bias parameter $\theta_3$ separated by semi colon (;). The smooth approximation of this program, to enable backpropagation, is shown in Eq~\ref{eq xor smooth}. 
\begin{equation}
\small
\label{eq xor smooth}
     \begin{aligned}
            &\mathtt{ \sigma(\beta\times affine_{[1,1;0]}(x))\times}\\
            &\mathtt{(\sigma(\beta\times affine_{[1,1;-1]}(x))\times affine_{[0,0;0]}(x)} + \\
            &\mathtt{(1-\sigma(\beta\times affine_{[1,1;-1]}(x)))\times affine_{[1,1;0]}(x))+}\\
            &\mathtt{(1-\sigma(\beta\times affine_{[1,1;0]}(x)))\times affine_{[0,0;0]}(x)}
        \end{aligned}
\end{equation}

In Eq~\ref{eq xor smooth}, $\sigma$ is the \textit{sigmoid} function. $\beta$ is a temperature parameter. As $\beta\rightarrow 0$, the approximation approaches usual $\mathtt{if-then-else}$ (\S~\ref{sec:dsl} of the main paper). The parameter values are hard-coded for illustration purposes. In practice, these weights are learned by training through gradient descent.

\section{Interpretability of Synthesized Programs: A Real World Example}
\label{appx_sec_interpretability}

We expect that each program primitive in a domain-specific language has a semantic meaning; hence, interpretability in program synthesis refers to understanding the decision of a synthesized program using various aspects such as: which program primitives are used and why? what does the learned sequence of program primitives mean for the problem? what is the effect of each program primitive on the output? etc.

\begin{figure}
    \centering
    \includegraphics[width=0.2\textwidth]{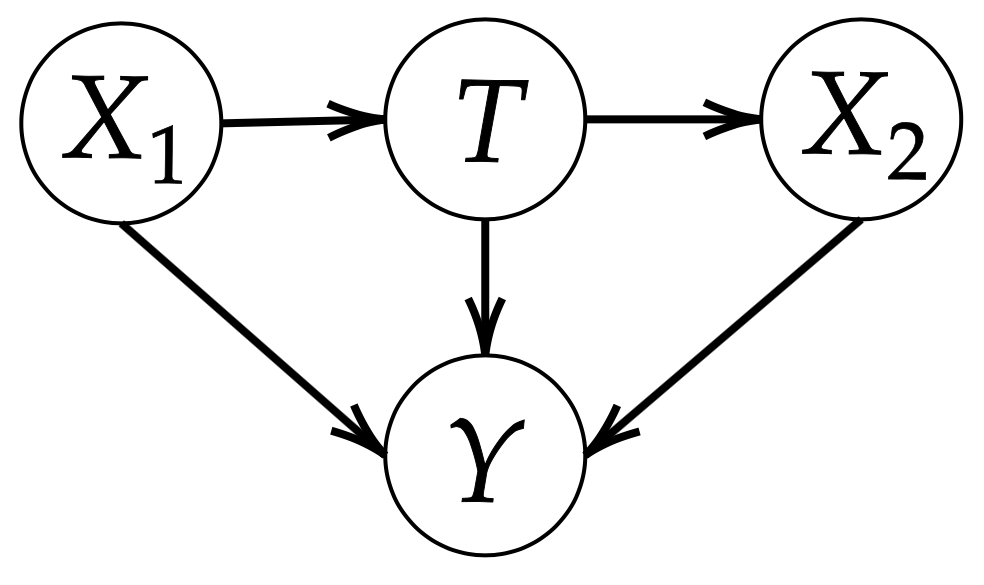}
    \caption{A real-world example for interpreting the synthesized programs.}
    \label{fig:realworldexample}
\end{figure}

We explain more clearly with an example. Consider a causal model consisting of variables $T,X_1,X_2,Y$ as shown in Fig~\ref{fig:realworldexample} where: (i) $X_1$ causes $T$ and $Y$; (ii) $T$ causes $X_2$ and $Y$; and (iii) $X_2$ causes $Y$. A real-world scenario depicted by this causal model could be where $T$ is the \textit{average distance walked by a person in a day}, $X_1$ is \textit{age}, $X_2$ is \textit{metabolism}, and $Y$ is \textit{blood pressure}. In this example, our goal is to estimate the effect of \textit{walking} ($T$) on \textit{blood pressure} ($Y$).  In this case, the ideal estimator for the quantity $\mathbb{E}[Y|do(t)]$ is $\sum_{x_1\sim X_1}\mathbb{E}[Y|t,x_1]p(x_1)$. However, NESTER has access to only observational data and is unaware of the underlying causal process. Now consider the following two possible programs $p_1, p_2$ synthesized by NESTER to estimate the causal effect of $T$ on $Y$. Let $\mathbf{v}=[t,x_1,x_2]$ be an input data point.
\begin{align*}
     p_1: \hspace{5pt}
    &\mathtt{if\hspace{4pt} mlp_0(subset(\mathbf{v},\{0\}))}
        \mathtt{\hspace{4pt}} \\ 
        &\mathtt{then \hspace{4pt}\ mlp_1(subset(\mathbf{v},\{0,1\}))}\\
        &\mathtt{\hspace{4pt} else\ \ mlp_2(subset(\mathbf{v},\{0,1\}))}
\end{align*}
\begin{align*}
    p_2: \hspace{5pt}
    &\mathtt{if\hspace{4pt} mlp_0(subset(\mathbf{v},\{0\}))}
        \mathtt{\hspace{4pt}} \\ 
        &\mathtt{then \hspace{4pt}\ mlp_1(subset(\mathbf{v},\{0,1,2\}))}\\
        &\mathtt{\hspace{4pt} else\ \ mlp_2(subset(\mathbf{v},\{0,1,2\}))}
\end{align*}
The only difference between $p_1$ and $p_2$ is the set of indices used in $\mathtt{subset}$ primitives. $p_1$ uses only $T, X_1$ (indicated by $\{0,1\}$ in $p_1$) to predict $Y$; while $p_2$ uses $T,X_1,X_2$ (indicated by $\{0,1,2\}$ in $p_2$) to predict $Y$. In this case, we would ideally observe $p_1$ to perform better than $p_2$ because $p_1$ controls for the correct set of confounding variables ($\{X_1\}$ in this case). Conversely, observing a strong performance for $p_1$ tells us that $\{X_1\}$ is the confounder, without knowledge of the causal model.
Observing the generated program and primitives gives us insights about the underlying data-generating process such as which features are the potential causes of treatment (e.g., \textit{age} affects the \textit{average distance} a person can walk), which features should not be controlled (e.g., we need the effect of \textit{walking} on \textit{blood pressure} irrespective of the \textit{metabolism} rate of a person), etc. Such information encoded in a synthesized program can also be validated with domain experts if available. Our experimental results and ablation studies discussed above show other ways of interpreting programs.

\section{Related Work}
\label{sec additional related work}
\noindent \textbf{Generative Modeling for Causal Effect Estimation:}
Generative Adversarial Networks (GANs)~\citep{goodfellow2014generative} have been used to learn the interventional distribution from observed data in both categorical and continuous treatment variable settings to estimate causal effects~\citep{yoon2018ganite,continuous_ganite}. By disentangling confounding variables from instrumental variables,~\citet{disent} proposed a variational inference method that uses only confounding variables to estimate causal effects. However, generative modeling requires a large amount of data to be useful, which is often not practical in causal effect estimation tasks.\citet{repr} proposed a method to learn representations by leveraging local similarities and thereby estimate causal effects. Ensemble models such as causal forests~\citep{causalforest}, and Bayesian additive regression trees~\citep{bart} have also been considered for causal effect estimation.

\noindent \textbf{Relevance of Causal Discovery Methods:}
In Pearlian approaches to causal effect estimation (which assume knowledge of causal graph), performing \textit{causal discovery} before causal effect estimation has also been studied in literature~\citep{nonlinearcd,mooij2016distinguishing,maathuis2010predicting,gupta2022local}. However, NN-based learning approaches have primarily focused on the potential outcomes approach for this objective. Since our work is situated in the latter context, under the \textit{ignorability} assumption (see \S~\ref{sec_background} of the main paper), we assume that the underlying causal graph takes a form in which the treatment is independent of the potential outcomes given a set (possibly empty) of pre-treatment covariates~\citep{tarnet}. Thus, we avoid performing causal discovery before estimating causal effects. 

\noindent \textbf{Neurosymbolic Program Synthesis (NPS) vs Neural Architecture Search (NAS):}
NAS, similar to NPS, is a technique to automatically design the best NN architecture to solve particular problems~\citep{nasrl, nasprogressive,nasps}. The significant difference between NPS and NAS is that, in NPS, symbolic and domain knowledge can be introduced in terms of the program primitives of a DSL. However, the goal in NAS is to design the best-performing architecture using a combination of standard neural network components, such as convolution blocks. In NPS, DSL changes for different applications, whereas in NAS, the underlying neural network blocks are fixed for all the problems. In NPS, we can combine symbolic reasoning and representation learning algorithms, making it a good choice for causal effect estimation.

%% file: images/cfr.tikz
\tikzset{every picture/.style={line width=0.75pt}} %set default line width to 0.75pt        

\begin{tikzpicture}[x=0.75pt,y=0.75pt,yscale=-1,xscale=1]

%Shape: Rectangle [id:dp18630427180201126] 
\draw  [color={rgb, 255:red, 0; green, 0; blue, 255 }  ,draw opacity=1 ][fill={rgb, 255:red, 255; green, 255; blue, 255 }  ,fill opacity=1 ][line width=1.5]  (5.33,75.17) -- (23.33,75.17) -- (23.33,157.17) -- (5.33,157.17) -- cycle ;
%Straight Lines [id:da46743664095742354] 
\draw [color={rgb, 255:red, 0; green, 0; blue, 255 }  ,draw opacity=1 ][line width=1.5]    (23.33,115.17) -- (42.34,115.99) ;
\draw [shift={(46.33,116.17)}, rotate = 182.49] [fill={rgb, 255:red, 0; green, 0; blue, 255 }  ,fill opacity=1 ][line width=0.08]  [draw opacity=0] (13.4,-6.43) -- (0,0) -- (13.4,6.44) -- (8.9,0) -- cycle    ;
%Straight Lines [id:da521497055966656] 
\draw [color={rgb, 255:red, 0; green, 0; blue, 255 }  ,draw opacity=1 ][line width=1.5]    (65.33,116.17) -- (84.34,116.99) ;
\draw [shift={(88.33,117.17)}, rotate = 182.49] [fill={rgb, 255:red, 0; green, 0; blue, 255 }  ,fill opacity=1 ][line width=0.08]  [draw opacity=0] (13.4,-6.43) -- (0,0) -- (13.4,6.44) -- (8.9,0) -- cycle    ;
%Shape: Rectangle [id:dp3018967017989128] 
\draw  [color={rgb, 255:red, 0; green, 0; blue, 255 }  ,draw opacity=1 ][fill={rgb, 255:red, 255; green, 255; blue, 255 }  ,fill opacity=1 ][line width=1.5]  (46.33,76.17) -- (64.33,76.17) -- (64.33,158.17) -- (46.33,158.17) -- cycle ;
%Shape: Rectangle [id:dp45935491355406466] 
\draw  [color={rgb, 255:red, 0; green, 0; blue, 255 }  ,draw opacity=1 ][fill={rgb, 255:red, 255; green, 255; blue, 255 }  ,fill opacity=1 ][line width=1.5]  (88.33,75.17) -- (106.33,75.17) -- (106.33,157.17) -- (88.33,157.17) -- cycle ;
%Shape: Rectangle [id:dp3914318931643095] 
\draw  [color={rgb, 255:red, 255; green, 0; blue, 255 }  ,draw opacity=1 ][fill={rgb, 255:red, 255; green, 255; blue, 255 }  ,fill opacity=1 ][line width=1.5]  (145.33,54.17) -- (163.33,54.17) -- (163.33,78.17) -- (145.33,78.17) -- cycle ;
%Shape: Rectangle [id:dp5170216308217981] 
\draw  [color={rgb, 255:red, 255; green, 0; blue, 0 }  ,draw opacity=1 ][fill={rgb, 255:red, 255; green, 255; blue, 255 }  ,fill opacity=1 ][line width=1.5]  (146.33,157.17) -- (164.33,157.17) -- (164.33,181.17) -- (146.33,181.17) -- cycle ;
%Straight Lines [id:da8633739879952721] 
\draw [color={rgb, 255:red, 150; green, 75; blue, 0 }  ,draw opacity=1 ][line width=1.5]    (107.33,116.17) -- (140.97,70.39) ;
\draw [shift={(143.33,67.17)}, rotate = 126.3] [fill={rgb, 255:red, 150; green, 75; blue, 0 }  ,fill opacity=1 ][line width=0.08]  [draw opacity=0] (13.4,-6.43) -- (0,0) -- (13.4,6.44) -- (8.9,0) -- cycle    ;
%Straight Lines [id:da7372353473945908] 
\draw [color={rgb, 255:red, 150; green, 75; blue, 0 }  ,draw opacity=1 ][line width=1.5]    (107.33,116.17) -- (141.09,165.86) ;
\draw [shift={(143.33,169.17)}, rotate = 235.81] [fill={rgb, 255:red, 150; green, 75; blue, 0 }  ,fill opacity=1 ][line width=0.08]  [draw opacity=0] (13.4,-6.43) -- (0,0) -- (13.4,6.44) -- (8.9,0) -- cycle    ;
%Shape: Rectangle [id:dp9453493546579506] 
\draw  [color={rgb, 255:red, 255; green, 0; blue, 255 }  ,draw opacity=1 ][fill={rgb, 255:red, 255; green, 255; blue, 255 }  ,fill opacity=1 ][line width=1.5]  (190.33,54.17) -- (208.33,54.17) -- (208.33,78.17) -- (190.33,78.17) -- cycle ;
%Straight Lines [id:da014892340875245491] 
\draw [color={rgb, 255:red, 255; green, 0; blue, 255 }  ,draw opacity=1 ][fill={rgb, 255:red, 255; green, 0; blue, 255 }  ,fill opacity=1 ][line width=1.5]    (163.33,66.17) -- (184.67,66.45) ;
\draw [shift={(188.67,66.5)}, rotate = 180.75] [fill={rgb, 255:red, 255; green, 0; blue, 255 }  ,fill opacity=1 ][line width=0.08]  [draw opacity=0] (13.4,-6.43) -- (0,0) -- (13.4,6.44) -- (8.9,0) -- cycle    ;
%Shape: Rectangle [id:dp26631806591847573] 
\draw  [color={rgb, 255:red, 255; green, 0; blue, 0 }  ,draw opacity=1 ][fill={rgb, 255:red, 255; green, 255; blue, 255 }  ,fill opacity=1 ][line width=1.5]  (191.33,157.17) -- (209.33,157.17) -- (209.33,181.17) -- (191.33,181.17) -- cycle ;
%Straight Lines [id:da10542891366339402] 
\draw [color={rgb, 255:red, 255; green, 0; blue, 0 }  ,draw opacity=1 ][line width=1.5]    (164.67,169.83) -- (187,170.4) ;
\draw [shift={(191,170.5)}, rotate = 181.45] [fill={rgb, 255:red, 255; green, 0; blue, 0 }  ,fill opacity=1 ][line width=0.08]  [draw opacity=0] (13.4,-6.43) -- (0,0) -- (13.4,6.44) -- (8.9,0) -- cycle    ;
%Shape: Rectangle [id:dp16065677340298223] 
\draw  [color={rgb, 255:red, 255; green, 0; blue, 255 }  ,draw opacity=1 ][fill={rgb, 255:red, 255; green, 255; blue, 255 }  ,fill opacity=1 ][line width=1.5]  (235.33,54.17) -- (253.33,54.17) -- (253.33,78.17) -- (235.33,78.17) -- cycle ;
%Straight Lines [id:da32934227264830573] 
\draw [color={rgb, 255:red, 255; green, 0; blue, 255 }  ,draw opacity=1 ][fill={rgb, 255:red, 255; green, 0; blue, 255 }  ,fill opacity=1 ][line width=1.5]    (209,67.17) -- (231,66.88) ;
\draw [shift={(235,66.83)}, rotate = 179.27] [fill={rgb, 255:red, 255; green, 0; blue, 255 }  ,fill opacity=1 ][line width=0.08]  [draw opacity=0] (13.4,-6.43) -- (0,0) -- (13.4,6.44) -- (8.9,0) -- cycle    ;
%Shape: Rectangle [id:dp3233359471543643] 
\draw  [color={rgb, 255:red, 255; green, 0; blue, 0 }  ,draw opacity=1 ][fill={rgb, 255:red, 255; green, 255; blue, 255 }  ,fill opacity=1 ][line width=1.5]  (236.33,158.17) -- (254.33,158.17) -- (254.33,182.17) -- (236.33,182.17) -- cycle ;
%Straight Lines [id:da6698805966926898] 
\draw [color={rgb, 255:red, 255; green, 0; blue, 0 }  ,draw opacity=1 ][line width=1.5]    (209.33,169.17) -- (231.67,169.45) ;
\draw [shift={(235.67,169.5)}, rotate = 180.73] [fill={rgb, 255:red, 255; green, 0; blue, 0 }  ,fill opacity=1 ][line width=0.08]  [draw opacity=0] (13.4,-6.43) -- (0,0) -- (13.4,6.44) -- (8.9,0) -- cycle    ;
%Straight Lines [id:da3931380001574123] 
\draw [color={rgb, 255:red, 0; green, 0; blue, 255 }  ,draw opacity=1 ][line width=1.5]    (96.67,158.67) -- (97.61,224.17) ;
\draw [shift={(97.67,228.17)}, rotate = 269.18] [fill={rgb, 255:red, 0; green, 0; blue, 255 }  ,fill opacity=1 ][line width=0.08]  [draw opacity=0] (13.4,-6.43) -- (0,0) -- (13.4,6.44) -- (8.9,0) -- cycle    ;
%Straight Lines [id:da7828730928145996] 
\draw [color={rgb, 255:red, 0; green, 0; blue, 255 }  ,draw opacity=1 ][line width=1.5]    (97.67,228.17) -- (142,228.47) ;
\draw [shift={(146,228.5)}, rotate = 180.4] [fill={rgb, 255:red, 0; green, 0; blue, 255 }  ,fill opacity=1 ][line width=0.08]  [draw opacity=0] (13.4,-6.43) -- (0,0) -- (13.4,6.44) -- (8.9,0) -- cycle    ;

% Text Node
\draw (89.33,252.07) node [anchor=north west][inner sep=0.75pt]  [font=\huge]  {$( a) \ CFR$};
% Text Node
\draw (292,73.4) node [anchor=north west][inner sep=0.75pt]  [font=\LARGE]  {$ \begin{array}{l}
\mathtt{\textcolor[rgb]{0.59,0.29,0}{if\ mlp}\textcolor[rgb]{0.59,0.29,0}{(}\textcolor[rgb]{0.59,0.29,0}{subset\ }\textcolor[rgb]{0.59,0.29,0}{(}\textcolor[rgb]{0.59,0.29,0}{[}\textcolor[rgb]{0.59,0.29,0}{t,X}\textcolor[rgb]{0.59,0.29,0}{]}\textcolor[rgb]{0.59,0.29,0}{,}\textcolor[rgb]{0.59,0.29,0}{\{}\textcolor[rgb]{0.59,0.29,0}{0}\textcolor[rgb]{0.59,0.29,0}{\}}\textcolor[rgb]{0.59,0.29,0}{)}\textcolor[rgb]{0.59,0.29,0}{)}\textcolor[rgb]{0.59,0.29,0}{ >0}}\\
\mathtt{\textcolor[rgb]{0.59,0.29,0}{then} \ \ \textcolor[rgb]{1,0,1}{mlp}\textcolor[rgb]{1,0,1}{_{1}}\textcolor[rgb]{1,0,1}{(}\textcolor[rgb]{0,0,1}{align}\textcolor[rgb]{0,0,1}{(}\textcolor[rgb]{0,0,1}{X}\textcolor[rgb]{0,0,1}{)}}\textcolor[rgb]{1,0,1}{)}\\
\mathtt{\textcolor[rgb]{0.59,0.29,0}{else} \ \ \textcolor[rgb]{1,0,0}{mlp}\textcolor[rgb]{1,0,0}{_{2}}\textcolor[rgb]{1,0,0}{(}\textcolor[rgb]{0,0,1}{align}\textcolor[rgb]{0,0,1}{(}\textcolor[rgb]{0,0,1}{X}}\textcolor[rgb]{0,0,1}{)}\textcolor[rgb]{1,0,0}{)}\\
\end{array}$};
% Text Node
\draw (432,243.4) node [anchor=north west][inner sep=0.75pt]  [font=\huge]  {$( b) \ \mathcal{P}_{C}$};
% Text Node
\draw (3.33,104.9) node [anchor=north west][inner sep=0.75pt]  [font=\huge,color={rgb, 255:red, 0; green, 0; blue, 255 }  ,opacity=1 ]  {$X$};
% Text Node
\draw (126,86.23) node [anchor=north west][inner sep=0.75pt]  [font=\huge,color={rgb, 255:red, 150; green, 75; blue, 0 }  ,opacity=1 ]  {$t=1$};
% Text Node
\draw (126.33,122.9) node [anchor=north west][inner sep=0.75pt]  [font=\huge,color={rgb, 255:red, 150; green, 75; blue, 0 }  ,opacity=1 ]  {$t=0$};
% Text Node
\draw (89.33,102.57) node [anchor=north west][inner sep=0.75pt]  [font=\huge,color={rgb, 255:red, 0; green, 0; blue, 255 }  ,opacity=1 ]  {$\phi $};
% Text Node
\draw (229.33,19.07) node [anchor=north west][inner sep=0.75pt]  [font=\huge,color={rgb, 255:red, 255; green, 0; blue, 255 }  ,opacity=1 ]  {$\hat{Y}_{1}$};
% Text Node
\draw (232.33,122.73) node [anchor=north west][inner sep=0.75pt]  [font=\huge,color={rgb, 255:red, 255; green, 0; blue, 0 }  ,opacity=1 ]  {$\hat{Y}_{0}$};
% Text Node
\draw (132,198.4) node [anchor=north west][inner sep=0.75pt]  [font=\huge,color={rgb, 255:red, 0; green, 0; blue, 255 }  ,opacity=1 ]  {$IPM( \phi ( X) |t=1,\phi ( X) |t=1)$};

\end{tikzpicture}

%% file: images/dragonnet.tikz
\tikzset{every picture/.style={line width=0.75pt}} %set default line width to 0.75pt        

\begin{tikzpicture}[x=0.75pt,y=0.75pt,yscale=-1,xscale=1]

%Shape: Rectangle [id:dp18630427180201126] 
\draw  [color={rgb, 255:red, 0; green, 0; blue, 255 }  ,draw opacity=1 ][fill={rgb, 255:red, 255; green, 255; blue, 255 }  ,fill opacity=1 ][line width=1.5]  (5.33,75.17) -- (23.33,75.17) -- (23.33,157.17) -- (5.33,157.17) -- cycle ;
%Straight Lines [id:da46743664095742354] 
\draw [color={rgb, 255:red, 0; green, 0; blue, 255 }  ,draw opacity=1 ][line width=1.5]    (23.33,115.17) -- (42.34,115.99) ;
\draw [shift={(46.33,116.17)}, rotate = 182.49] [fill={rgb, 255:red, 0; green, 0; blue, 255 }  ,fill opacity=1 ][line width=0.08]  [draw opacity=0] (13.4,-6.43) -- (0,0) -- (13.4,6.44) -- (8.9,0) -- cycle    ;
%Straight Lines [id:da521497055966656] 
\draw [color={rgb, 255:red, 0; green, 0; blue, 255 }  ,draw opacity=1 ][line width=1.5]    (65.33,116.17) -- (84.34,116.99) ;
\draw [shift={(88.33,117.17)}, rotate = 182.49] [fill={rgb, 255:red, 0; green, 0; blue, 255 }  ,fill opacity=1 ][line width=0.08]  [draw opacity=0] (13.4,-6.43) -- (0,0) -- (13.4,6.44) -- (8.9,0) -- cycle    ;
%Shape: Rectangle [id:dp3018967017989128] 
\draw  [color={rgb, 255:red, 0; green, 0; blue, 255 }  ,draw opacity=1 ][fill={rgb, 255:red, 255; green, 255; blue, 255 }  ,fill opacity=1 ][line width=1.5]  (46.33,76.17) -- (64.33,76.17) -- (64.33,158.17) -- (46.33,158.17) -- cycle ;
%Shape: Rectangle [id:dp45935491355406466] 
\draw  [color={rgb, 255:red, 0; green, 0; blue, 255 }  ,draw opacity=1 ][fill={rgb, 255:red, 255; green, 255; blue, 255 }  ,fill opacity=1 ][line width=1.5]  (88.33,75.17) -- (106.33,75.17) -- (106.33,157.17) -- (88.33,157.17) -- cycle ;
%Shape: Rectangle [id:dp3914318931643095] 
\draw  [color={rgb, 255:red, 255; green, 0; blue, 255 }  ,draw opacity=1 ][fill={rgb, 255:red, 255; green, 255; blue, 255 }  ,fill opacity=1 ][line width=1.5]  (145.33,54.17) -- (163.33,54.17) -- (163.33,78.17) -- (145.33,78.17) -- cycle ;
%Shape: Rectangle [id:dp5170216308217981] 
\draw  [color={rgb, 255:red, 255; green, 0; blue, 0 }  ,draw opacity=1 ][fill={rgb, 255:red, 255; green, 255; blue, 255 }  ,fill opacity=1 ][line width=1.5]  (146.33,157.17) -- (164.33,157.17) -- (164.33,181.17) -- (146.33,181.17) -- cycle ;
%Straight Lines [id:da8633739879952721] 
\draw [color={rgb, 255:red, 150; green, 75; blue, 0 }  ,draw opacity=1 ][line width=1.5]    (107.33,116.17) -- (140.97,70.39) ;
\draw [shift={(143.33,67.17)}, rotate = 126.3] [fill={rgb, 255:red, 150; green, 75; blue, 0 }  ,fill opacity=1 ][line width=0.08]  [draw opacity=0] (13.4,-6.43) -- (0,0) -- (13.4,6.44) -- (8.9,0) -- cycle    ;
%Straight Lines [id:da7372353473945908] 
\draw [color={rgb, 255:red, 150; green, 75; blue, 0 }  ,draw opacity=1 ][line width=1.5]    (107.33,116.17) -- (141.09,165.86) ;
\draw [shift={(143.33,169.17)}, rotate = 235.81] [fill={rgb, 255:red, 150; green, 75; blue, 0 }  ,fill opacity=1 ][line width=0.08]  [draw opacity=0] (13.4,-6.43) -- (0,0) -- (13.4,6.44) -- (8.9,0) -- cycle    ;
%Shape: Rectangle [id:dp9453493546579506] 
\draw  [color={rgb, 255:red, 255; green, 0; blue, 255 }  ,draw opacity=1 ][fill={rgb, 255:red, 255; green, 255; blue, 255 }  ,fill opacity=1 ][line width=1.5]  (190.33,54.17) -- (208.33,54.17) -- (208.33,78.17) -- (190.33,78.17) -- cycle ;
%Straight Lines [id:da014892340875245491] 
\draw [color={rgb, 255:red, 255; green, 0; blue, 255 }  ,draw opacity=1 ][fill={rgb, 255:red, 255; green, 0; blue, 255 }  ,fill opacity=1 ][line width=1.5]    (163.33,66.17) -- (184.67,66.45) ;
\draw [shift={(188.67,66.5)}, rotate = 180.75] [fill={rgb, 255:red, 255; green, 0; blue, 255 }  ,fill opacity=1 ][line width=0.08]  [draw opacity=0] (13.4,-6.43) -- (0,0) -- (13.4,6.44) -- (8.9,0) -- cycle    ;
%Shape: Rectangle [id:dp26631806591847573] 
\draw  [color={rgb, 255:red, 255; green, 0; blue, 0 }  ,draw opacity=1 ][fill={rgb, 255:red, 255; green, 255; blue, 255 }  ,fill opacity=1 ][line width=1.5]  (191.33,157.17) -- (209.33,157.17) -- (209.33,181.17) -- (191.33,181.17) -- cycle ;
%Straight Lines [id:da10542891366339402] 
\draw [color={rgb, 255:red, 255; green, 0; blue, 0 }  ,draw opacity=1 ][line width=1.5]    (164.67,169.83) -- (187,170.4) ;
\draw [shift={(191,170.5)}, rotate = 181.45] [fill={rgb, 255:red, 255; green, 0; blue, 0 }  ,fill opacity=1 ][line width=0.08]  [draw opacity=0] (13.4,-6.43) -- (0,0) -- (13.4,6.44) -- (8.9,0) -- cycle    ;
%Shape: Rectangle [id:dp16065677340298223] 
\draw  [color={rgb, 255:red, 255; green, 0; blue, 255 }  ,draw opacity=1 ][fill={rgb, 255:red, 255; green, 255; blue, 255 }  ,fill opacity=1 ][line width=1.5]  (235.33,54.17) -- (253.33,54.17) -- (253.33,78.17) -- (235.33,78.17) -- cycle ;
%Straight Lines [id:da32934227264830573] 
\draw [color={rgb, 255:red, 255; green, 0; blue, 255 }  ,draw opacity=1 ][fill={rgb, 255:red, 255; green, 0; blue, 255 }  ,fill opacity=1 ][line width=1.5]    (209,67.17) -- (231,66.88) ;
\draw [shift={(235,66.83)}, rotate = 179.27] [fill={rgb, 255:red, 255; green, 0; blue, 255 }  ,fill opacity=1 ][line width=0.08]  [draw opacity=0] (13.4,-6.43) -- (0,0) -- (13.4,6.44) -- (8.9,0) -- cycle    ;
%Shape: Rectangle [id:dp3233359471543643] 
\draw  [color={rgb, 255:red, 255; green, 0; blue, 0 }  ,draw opacity=1 ][fill={rgb, 255:red, 255; green, 255; blue, 255 }  ,fill opacity=1 ][line width=1.5]  (236.33,158.17) -- (254.33,158.17) -- (254.33,182.17) -- (236.33,182.17) -- cycle ;
%Straight Lines [id:da6698805966926898] 
\draw [color={rgb, 255:red, 255; green, 0; blue, 0 }  ,draw opacity=1 ][line width=1.5]    (209.33,169.17) -- (231.67,169.45) ;
\draw [shift={(235.67,169.5)}, rotate = 180.73] [fill={rgb, 255:red, 255; green, 0; blue, 0 }  ,fill opacity=1 ][line width=0.08]  [draw opacity=0] (13.4,-6.43) -- (0,0) -- (13.4,6.44) -- (8.9,0) -- cycle    ;
%Straight Lines [id:da3931380001574123] 
\draw [color={rgb, 255:red, 0; green, 0; blue, 255 }  ,draw opacity=1 ][line width=1.5]    (97,157.5) -- (142.8,226.99) ;
\draw [shift={(145,230.33)}, rotate = 236.61] [fill={rgb, 255:red, 0; green, 0; blue, 255 }  ,fill opacity=1 ][line width=0.08]  [draw opacity=0] (13.4,-6.43) -- (0,0) -- (13.4,6.44) -- (8.9,0) -- cycle    ;
%Shape: Rectangle [id:dp43609705795303255] 
\draw  [color={rgb, 255:red, 0; green, 0; blue, 255 }  ,draw opacity=1 ][fill={rgb, 255:red, 255; green, 255; blue, 255 }  ,fill opacity=1 ][line width=1.5]  (146.33,215.5) -- (164.33,215.5) -- (164.33,239.5) -- (146.33,239.5) -- cycle ;
%Shape: Rectangle [id:dp8242172650940928] 
\draw  [color={rgb, 255:red, 0; green, 0; blue, 255 }  ,draw opacity=1 ][fill={rgb, 255:red, 255; green, 255; blue, 255 }  ,fill opacity=1 ][line width=1.5]  (191.33,215.5) -- (209.33,215.5) -- (209.33,239.5) -- (191.33,239.5) -- cycle ;
%Straight Lines [id:da19828115959213832] 
\draw [color={rgb, 255:red, 0; green, 0; blue, 255 }  ,draw opacity=1 ][line width=1.5]    (164.67,228.17) -- (187,228.73) ;
\draw [shift={(191,228.83)}, rotate = 181.45] [fill={rgb, 255:red, 0; green, 0; blue, 255 }  ,fill opacity=1 ][line width=0.08]  [draw opacity=0] (13.4,-6.43) -- (0,0) -- (13.4,6.44) -- (8.9,0) -- cycle    ;
%Shape: Rectangle [id:dp8530668696436314] 
\draw  [color={rgb, 255:red, 0; green, 0; blue, 255 }  ,draw opacity=1 ][fill={rgb, 255:red, 255; green, 255; blue, 255 }  ,fill opacity=1 ][line width=1.5]  (236.33,216.5) -- (254.33,216.5) -- (254.33,240.5) -- (236.33,240.5) -- cycle ;
%Straight Lines [id:da03534115628490342] 
\draw [color={rgb, 255:red, 0; green, 0; blue, 255 }  ,draw opacity=1 ][line width=1.5]    (210.67,227.17) -- (233,227.73) ;
\draw [shift={(237,227.83)}, rotate = 181.45] [fill={rgb, 255:red, 0; green, 0; blue, 255 }  ,fill opacity=1 ][line width=0.08]  [draw opacity=0] (13.4,-6.43) -- (0,0) -- (13.4,6.44) -- (8.9,0) -- cycle    ;

% Text Node
\draw (89.33,252.07) node [anchor=north west][inner sep=0.75pt]  [font=\huge]  {$( a) \ Dragonnet$};
% Text Node
\draw (299.67,66.87) node [anchor=north west][inner sep=0.75pt]  [font=\LARGE]  {$ \begin{array}{l}
\mathtt{\textcolor[rgb]{0.59,0.29,0}{if\ mlp}\textcolor[rgb]{0.59,0.29,0}{(}\textcolor[rgb]{0.59,0.29,0}{subset\ }\textcolor[rgb]{0.59,0.29,0}{(}\textcolor[rgb]{0.59,0.29,0}{[}\textcolor[rgb]{0.59,0.29,0}{t,X}\textcolor[rgb]{0.59,0.29,0}{]}\textcolor[rgb]{0.59,0.29,0}{,}\textcolor[rgb]{0.59,0.29,0}{\{}\textcolor[rgb]{0.59,0.29,0}{0}\textcolor[rgb]{0.59,0.29,0}{\}}\textcolor[rgb]{0.59,0.29,0}{)}\textcolor[rgb]{0.59,0.29,0}{)}\textcolor[rgb]{0.59,0.29,0}{ >0}}\\
\mathtt{\textcolor[rgb]{0.59,0.29,0}{then} \ \ \textcolor[rgb]{1,0,1}{mlp}\textcolor[rgb]{1,0,1}{_{1}}\textcolor[rgb]{1,0,1}{(}\textcolor[rgb]{0,0,1}{propensity}\textcolor[rgb]{0,0,1}{(}\textcolor[rgb]{0,0,1}{X}\textcolor[rgb]{0,0,1}{)}}\textcolor[rgb]{1,0,1}{)}\\
\mathtt{\textcolor[rgb]{0.59,0.29,0}{else} \ \ \textcolor[rgb]{1,0,0}{mlp}\textcolor[rgb]{1,0,0}{_{2}}\textcolor[rgb]{1,0,0}{(}\textcolor[rgb]{0,0,1}{propensity}\textcolor[rgb]{0,0,1}{(}\textcolor[rgb]{0,0,1}{X}}\textcolor[rgb]{0,0,1}{)}\textcolor[rgb]{1,0,0}{)}\\
\end{array}$};
% Text Node
\draw (422,254.07) node [anchor=north west][inner sep=0.75pt]  [font=\huge]  {$( b) \ \mathcal{P}_{D}$};
% Text Node
\draw (3.33,104.9) node [anchor=north west][inner sep=0.75pt]  [font=\huge,color={rgb, 255:red, 0; green, 0; blue, 255 }  ,opacity=1 ]  {$X$};
% Text Node
\draw (126,86.23) node [anchor=north west][inner sep=0.75pt]  [font=\huge,color={rgb, 255:red, 150; green, 75; blue, 0 }  ,opacity=1 ]  {$t=1$};
% Text Node
\draw (126.33,122.9) node [anchor=north west][inner sep=0.75pt]  [font=\huge,color={rgb, 255:red, 150; green, 75; blue, 0 }  ,opacity=1 ]  {$t=0$};
% Text Node
\draw (89.33,102.57) node [anchor=north west][inner sep=0.75pt]  [font=\huge,color={rgb, 255:red, 0; green, 0; blue, 255 }  ,opacity=1 ]  {$\phi $};
% Text Node
\draw (229.33,19.07) node [anchor=north west][inner sep=0.75pt]  [font=\huge,color={rgb, 255:red, 255; green, 0; blue, 255 }  ,opacity=1 ]  {$\hat{Y}_{1}$};
% Text Node
\draw (232.33,122.73) node [anchor=north west][inner sep=0.75pt]  [font=\huge,color={rgb, 255:red, 255; green, 0; blue, 0 }  ,opacity=1 ]  {$\hat{Y}_{0}$};
% Text Node
\draw (235.33,186.07) node [anchor=north west][inner sep=0.75pt]  [font=\huge,color={rgb, 255:red, 0; green, 0; blue, 255 }  ,opacity=1 ]  {$\hat{t}$};

\end{tikzpicture}

%% file: main.bbl
\begin{thebibliography}{67}
\providecommand{\natexlab}[1]{#1}

\bibitem[{A.~Smith and E.~Todd(2005)}]{jobs1}
A.~Smith, J.; and E.~Todd, P. 2005.
\newblock {Does matching overcome LaLonde's critique of nonexperimental
  estimators?}
\newblock \emph{Journal of Econometrics}, 125(1-2): 305--353.

\bibitem[{Abadie and Imbens(2006)}]{matching}
Abadie, A.; and Imbens, G.~W. 2006.
\newblock Large sample properties of matching estimators for average treatment
  effects.
\newblock \emph{Econometrica}, 74(1): 235--267.

\bibitem[{Almond, Chay, and Lee(2005)}]{twins}
Almond, D.; Chay, K.~Y.; and Lee, D.~S. 2005.
\newblock The Costs of Low Birth Weight.
\newblock \emph{The Quarterly Journal of Economics}, 120(3): 1031--1083.

\bibitem[{Assaad et~al.(2021)Assaad, Zeng, Tao, Datta, Mehta, Henao, Li, and
  Carin~Duke}]{balancing}
Assaad, S.; Zeng, S.; Tao, C.; Datta, S.; Mehta, N.; Henao, R.; Li, F.; and
  Carin~Duke, L. 2021.
\newblock Counterfactual Representation Learning with Balancing Weights.
\newblock In \emph{AISTATS}.

\bibitem[{Bica, Jordon, and van~der Schaar(2020)}]{continuous_ganite}
Bica, I.; Jordon, J.; and van~der Schaar, M. 2020.
\newblock Estimating the Effects of Continuous-valued Interventions using
  Generative Adversarial Networks.
\newblock In \emph{NeurIPS}.

\bibitem[{Biermann(1978)}]{Biermmann}
Biermann, A.~W. 1978.
\newblock The Inference of Regular LISP Programs from Examples.
\newblock \emph{IEEE Transactions on Systems, Man, and Cybernetics}, 8(8):
  585--600.

\bibitem[{Bo{\v{s}}njak et~al.(2017)Bo{\v{s}}njak, Rockt{\"a}schel, Naradowsky,
  and Riedel}]{bosnjak17a}
Bo{\v{s}}njak, M.; Rockt{\"a}schel, T.; Naradowsky, J.; and Riedel, S. 2017.
\newblock Programming with a Differentiable Forth Interpreter.
\newblock In \emph{ICML}.

\bibitem[{Brady, Collier, and Sekhon(2008)}]{neyman-rubin}
Brady, H.; Collier, D.; and Sekhon, J. 2008.
\newblock The Neyman-Rubin Model of Causal Inference and Estimation Via
  Matching Methods.
\newblock \emph{The Oxford Handbook of Political Methodology}.

\bibitem[{Carey and Stiles(2016)}]{carey2016some}
Carey, T.~A.; and Stiles, W.~B. 2016.
\newblock Some problems with randomized controlled trials and some viable
  alternatives.
\newblock \emph{Clinical Psychology \& Psychotherapy}, 23(1): 87--95.

\bibitem[{Chalmers et~al.(1981)Chalmers, Smith, Blackburn, Silverman,
  Schroeder, Reitman, and Ambroz}]{rct}
Chalmers, T.~C.; Smith, H.; Blackburn, B.; Silverman, B.; Schroeder, B.;
  Reitman, D.; and Ambroz, A. 1981.
\newblock A method for assessing the quality of a randomized control trial.
\newblock \emph{Controlled Clinical Trials}, 2(1): 31--49.

\bibitem[{Chipman, George, and McCulloch(2010)}]{bart}
Chipman, H.~A.; George, E.~I.; and McCulloch, R.~E. 2010.
\newblock {BART: Bayesian additive regression trees}.
\newblock \emph{The Annals of Applied Statistics}, 4(1): 266 -- 298.

\bibitem[{Chu, Rathbun, and Li(2020)}]{chu_balanced}
Chu, Z.; Rathbun, S.~L.; and Li, S. 2020.
\newblock Matching in Selective and Balanced Representation Space for Treatment
  Effects Estimation.
\newblock In \emph{CIKM}.

\bibitem[{Cinelli, Forney, and Pearl(2022)}]{crashcourse}
Cinelli, C.; Forney, A.; and Pearl, J. 2022.
\newblock A Crash Course in Good and Bad Controls.
\newblock \emph{Sociological Methods \& Research}.

\bibitem[{Collier and Mahoney(1996)}]{collier1996insights}
Collier, D.; and Mahoney, J. 1996.
\newblock Insights and pitfalls: Selection bias in qualitative research.
\newblock \emph{World politics}, 49(1): 56--91.

\bibitem[{CRUMP et~al.(2009)CRUMP, HOTZ, IMBENS, and MITNIK}]{overlap}
CRUMP, R.~K.; HOTZ, V.~J.; IMBENS, G.~W.; and MITNIK, O.~A. 2009.
\newblock Dealing with limited overlap in estimation of average treatment
  effects.
\newblock \emph{Biometrika}, 96(1): 187--199.

\bibitem[{Cui and Zhu(2021)}]{dpads}
Cui, G.; and Zhu, H. 2021.
\newblock Differentiable Synthesis of Program Architectures.
\newblock In \emph{NeurIPS}.

\bibitem[{Curth and van~der
  Schaar(2021{\natexlab{a}})}]{curth2021nonparametric}
Curth, A.; and van~der Schaar, M. 2021{\natexlab{a}}.
\newblock Nonparametric Estimation of Heterogeneous Treatment Effects: From
  Theory to Learning Algorithms.
\newblock In \emph{Proceedings of the 24th International Conference on
  Artificial Intelligence and Statistics (AISTATS)}. PMLR.

\bibitem[{Curth and van~der Schaar(2021{\natexlab{b}})}]{curth2021on}
Curth, A.; and van~der Schaar, M. 2021{\natexlab{b}}.
\newblock On Inductive Biases for Heterogeneous Treatment Effect Estimation.
\newblock In \emph{Advances in Neural Information Processing Systems}.

\bibitem[{Cuturi and Doucet(2014)}]{cuturi14}
Cuturi, M.; and Doucet, A. 2014.
\newblock Fast Computation of Wasserstein Barycenters.
\newblock In \emph{ICML}.

\bibitem[{Diamond and Sekhon(2013)}]{genetic}
Diamond, A.; and Sekhon, J.~S. 2013.
\newblock Genetic matching for estimating causal effects: A general
  multi-variate matching method for achieving balance in observational studies.
\newblock \emph{Review of Economics and Statistics}, 95(3): 932--945.

\bibitem[{Dorie(2016)}]{npci}
Dorie, V. 2016.
\newblock NPCI: Non-parametrics for Causal Inference.

\bibitem[{Farajtabar et~al.(2020)Farajtabar, Lee, Feng, Gupta, Dolan, Chandran,
  and Szummer}]{balance_reg}
Farajtabar, M.; Lee, A.; Feng, Y.; Gupta, V.; Dolan, P.; Chandran, H.; and
  Szummer, M. 2020.
\newblock Balance Regularized Neural Network Models for Causal Effect
  Estimation.
\newblock \emph{CoRR}, abs/2011.11199.

\bibitem[{Gaunt et~al.(2017)Gaunt, Brockschmidt, Kushman, and
  Tarlow}]{gaunt17a}
Gaunt, A.~L.; Brockschmidt, M.; Kushman, N.; and Tarlow, D. 2017.
\newblock Differentiable Programs with Neural Libraries.
\newblock In \emph{ICML}.

\bibitem[{Goodfellow et~al.(2014)Goodfellow, Pouget-Abadie, Mirza, Xu,
  Warde-Farley, Ozair, Courville, and Bengio}]{goodfellow2014generative}
Goodfellow, I.~J.; Pouget-Abadie, J.; Mirza, M.; Xu, B.; Warde-Farley, D.;
  Ozair, S.; Courville, A.~C.; and Bengio, Y. 2014.
\newblock Generative Adversarial Nets.
\newblock In \emph{NIPS}.

\bibitem[{Gretton et~al.(2012)Gretton, Borgwardt, Rasch, Sch{{\"o}}lkopf, and
  Smola}]{gretton12}
Gretton, A.; Borgwardt, K.~M.; Rasch, M.~J.; Sch{{\"o}}lkopf, B.; and Smola, A.
  2012.
\newblock A Kernel Two-Sample Test.
\newblock \emph{JMLR}, 13(25): 723--773.

\bibitem[{Gulwani(2011)}]{gulwani2011automating}
Gulwani, S. 2011.
\newblock Automating string processing in spreadsheets using input-output
  examples.
\newblock \emph{ACM Sigplan Notices}, 46(1): 317--330.

\bibitem[{Gupta, Childers, and Lipton(2022)}]{gupta2022local}
Gupta, S.; Childers, D.; and Lipton, Z.~C. 2022.
\newblock Local Causal Discovery for Estimating Causal Effects.
\newblock In \emph{NeurIPS 2022 Workshop on Causality for Real-world Impact}.

\bibitem[{Harris(1974)}]{harris}
Harris, L.~R. 1974.
\newblock The heuristic search under conditions of error.
\newblock \emph{Artificial Intelligence}, 5(3): 217--234.

\bibitem[{Hill(2011)}]{ihdp_dataset}
Hill, J.~L. 2011.
\newblock Bayesian Nonparametric Modeling for Causal Inference.
\newblock \emph{Journal of Computational and Graphical Statistics}, 20(1):
  217--240.

\bibitem[{Hopcroft, Motwani, and Ullman(2001)}]{ullman}
Hopcroft, J.~E.; Motwani, R.; and Ullman, J.~D. 2001.
\newblock Introduction to automata theory, languages, and computation.
\newblock \emph{Acm Sigact News}, 32(1): 60--65.

\bibitem[{Hornik, Stinchcombe, and White(1989)}]{hornik}
Hornik, K.; Stinchcombe, M.; and White, H. 1989.
\newblock Multilayer feedforward networks are universal approximators.
\newblock \emph{Neural networks}, 2(5): 359--366.

\bibitem[{Hoyer et~al.(2008)Hoyer, Janzing, Mooij, Peters, and
  Sch\"{o}lkopf}]{nonlinearcd}
Hoyer, P.; Janzing, D.; Mooij, J.~M.; Peters, J.; and Sch\"{o}lkopf, B. 2008.
\newblock Nonlinear causal discovery with additive noise models.
\newblock In \emph{Advances in Neural Information Processing Systems}.

\bibitem[{Imbens(2000)}]{imbens2000role}
Imbens, G.~W. 2000.
\newblock The role of the propensity score in estimating dose-response
  functions.
\newblock \emph{Biometrika}, 87(3): 706--710.

\bibitem[{Kocsis and Szepesv{\'a}ri(2006)}]{kocsis2006bandit}
Kocsis, L.; and Szepesv{\'a}ri, C. 2006.
\newblock Bandit based monte-carlo planning.
\newblock In \emph{European conference on machine learning}, 282--293.
  Springer.

\bibitem[{K{\"u}nzel et~al.(2019)K{\"u}nzel, Sekhon, Bickel, and Yu}]{xlearner}
K{\"u}nzel, S.~R.; Sekhon, J.~S.; Bickel, P.~J.; and Yu, B. 2019.
\newblock Metalearners for estimating heterogeneous treatment effects using
  machine learning.
\newblock \emph{Proceedings of the National Academy of Sciences}, 116(10):
  4156--4165.

\bibitem[{LaLonde(1986)}]{jobs}
LaLonde, R.~J. 1986.
\newblock Evaluating the Econometric Evaluations of Training Programs with
  Experimental Data.
\newblock \emph{The American Economic Review}, 76(4): 604--620.

\bibitem[{Lechner(2001)}]{lechner2001identification}
Lechner, M. 2001.
\newblock Identification and estimation of causal effects of multiple
  treatments under the conditional independence assumption.
\newblock In \emph{Econometric evaluation of labour market policies}, 43--58.
  Springer.

\bibitem[{Li and Fu(2017)}]{matchingon}
Li, S.; and Fu, Y. 2017.
\newblock Matching on Balanced Nonlinear Representations for Treatment Effects
  Estimation.
\newblock In Guyon, I.; Luxburg, U.~V.; Bengio, S.; Wallach, H.; Fergus, R.;
  Vishwanathan, S.; and Garnett, R., eds., \emph{NIPS}.

\bibitem[{Liu et~al.(2018)Liu, Zoph, Neumann, Shlens, Hua, Li, Fei-Fei, Yuille,
  Huang, and Murphy}]{nasprogressive}
Liu, C.; Zoph, B.; Neumann, M.; Shlens, J.; Hua, W.; Li, L.-J.; Fei-Fei, L.;
  Yuille, A.; Huang, J.; and Murphy, K. 2018.
\newblock Progressive Neural Architecture Search.
\newblock In \emph{ECCV}.

\bibitem[{Maathuis et~al.(2010)Maathuis, Colombo, Kalisch, and
  B{\"u}hlmann}]{maathuis2010predicting}
Maathuis, M.~H.; Colombo, D.; Kalisch, M.; and B{\"u}hlmann, P. 2010.
\newblock Predicting causal effects in large-scale systems from observational
  data.
\newblock \emph{Nature methods}, 7(4): 247--248.

\bibitem[{Mooij et~al.(2016)Mooij, Peters, Janzing, Zscheischler, and
  Sch{\"o}lkopf}]{mooij2016distinguishing}
Mooij, J.~M.; Peters, J.; Janzing, D.; Zscheischler, J.; and Sch{\"o}lkopf, B.
  2016.
\newblock Distinguishing cause from effect using observational data: methods
  and benchmarks.
\newblock \emph{The Journal of Machine Learning Research}, 17(1): 1103--1204.

\bibitem[{Morgan and Winship(2014)}]{morgan_winship_2014}
Morgan, S.~L.; and Winship, C. 2014.
\newblock \emph{Counterfactuals and Causal Inference: Methods and Principles
  for Social Research}.
\newblock Analytical Methods for Social Research. Cambridge University Press, 2
  edition.

\bibitem[{Parisotto et~al.(2017)Parisotto, rahman Mohamed, Singh, Li, Zhou, and
  Kohli}]{parisotto2016neuro}
Parisotto, E.; rahman Mohamed, A.; Singh, R.; Li, L.; Zhou, D.; and Kohli, P.
  2017.
\newblock Neuro-Symbolic Program Synthesis.
\newblock In \emph{ICLR}.

\bibitem[{Pearl(1984)}]{heuristics}
Pearl, J. 1984.
\newblock \emph{Heuristics: Intelligent Search Strategies for Computer Problem
  Solving}.
\newblock Addison-Wesley Longman Publishing Co., Inc.

\bibitem[{Pearl(2009)}]{pearl2009causality}
Pearl, J. 2009.
\newblock \emph{Causality}.
\newblock Cambridge university press.

\bibitem[{Pearl, Glymour, and Jewell(2016)}]{pearl2016causal}
Pearl, J.; Glymour, M.; and Jewell, N. 2016.
\newblock \emph{Causal Inference in Statistics: A Primer}.
\newblock Wiley.

\bibitem[{Pham et~al.(2018)Pham, Guan, Zoph, Le, and Dean}]{nasps}
Pham, H.; Guan, M.; Zoph, B.; Le, Q.; and Dean, J. 2018.
\newblock Efficient Neural Architecture Search via Parameters Sharing.
\newblock In \emph{ICML}.

\bibitem[{Rosenbaum and Rubin(1983)}]{rosenbaum1983central}
Rosenbaum, P.~R.; and Rubin, D.~B. 1983.
\newblock The central role of the propensity score in observational studies for
  causal effects.
\newblock \emph{Biometrika}, 70(1): 41--55.

\bibitem[{Rosenbaum and Rubin(1985)}]{rosenbaum1985constructing}
Rosenbaum, P.~R.; and Rubin, D.~B. 1985.
\newblock Constructing a control group using multivariate matched sampling
  methods that incorporate the propensity score.
\newblock \emph{The American Statistician}, 39(1): 33--38.

\bibitem[{Rubin(1974)}]{po}
Rubin, D.~B. 1974.
\newblock Estimating causal effects of treatments in randomized and
  nonrandomized studies.
\newblock \emph{Journal of educational Psychology}, 66(5): 688.

\bibitem[{Sanson-Fisher et~al.(2007)Sanson-Fisher, Bonevski, Green, and
  D’Este}]{rct_cost}
Sanson-Fisher, R.~W.; Bonevski, B.; Green, L.~W.; and D’Este, C. 2007.
\newblock Limitations of the Randomized Controlled Trial in Evaluating
  Population-Based Health Interventions.
\newblock \emph{American Journal of Preventive Medicine}, 33(2): 155--161.

\bibitem[{Schwab et~al.(2020)Schwab, Linhardt, Bauer, Buhmann, and
  Karlen}]{dr_net}
Schwab, P.; Linhardt, L.; Bauer, S.; Buhmann, J.~M.; and Karlen, W. 2020.
\newblock Learning counterfactual representations for estimating individual
  dose-response curves.
\newblock In \emph{AAAI}.

\bibitem[{Shah et~al.(2020)Shah, Zhan, Sun, Verma, Yue, and Chaudhuri}]{near}
Shah, A.; Zhan, E.; Sun, J.; Verma, A.; Yue, Y.; and Chaudhuri, S. 2020.
\newblock Learning Differentiable Programs with Admissible Neural Heuristics.
\newblock In \emph{NeurIPS}.

\bibitem[{Shalit, Johansson, and Sontag(2017)}]{tarnet}
Shalit, U.; Johansson, F.~D.; and Sontag, D. 2017.
\newblock Estimating individual treatment effect: generalization bounds and
  algorithms.
\newblock In \emph{ICML}.

\bibitem[{Shi, Blei, and Veitch(2019)}]{dragon_net}
Shi, C.; Blei, D.; and Veitch, V. 2019.
\newblock Adapting Neural Networks for the Estimation of Treatment Effects.
\newblock In \emph{NeurIPS}.

\bibitem[{Shi et~al.(2022)Shi, Dai, Ellis, and Sutton}]{crossbeam}
Shi, K.; Dai, H.; Ellis, K.; and Sutton, C. 2022.
\newblock CrossBeam: Learning to Search in Bottom-Up Program Synthesis.
\newblock In \emph{ICLR}.

\bibitem[{Solar~Lezama(2008)}]{Lezama}
Solar~Lezama, A. 2008.
\newblock \emph{Program Synthesis By Sketching}.
\newblock Ph.D. thesis, EECS Department, University of California, Berkeley.

\bibitem[{Solar-Lezama et~al.(2005)Solar-Lezama, Rabbah, Bod\'{\i}k, and
  Ebcio\u{g}lu}]{solar}
Solar-Lezama, A.; Rabbah, R.; Bod\'{\i}k, R.; and Ebcio\u{g}lu, K. 2005.
\newblock Programming by Sketching for Bit-Streaming Programs.
\newblock \emph{SIGPLAN Not.}, 40(6): 281–294.

\bibitem[{Stuart(2010)}]{stuart2010matching}
Stuart, E.~A. 2010.
\newblock Matching methods for causal inference: A review and a look forward.
\newblock \emph{Statistical science: a review journal of the Institute of
  Mathematical Statistics}, 25(1): 1.

\bibitem[{Tang and Ellis(2022)}]{percepreason}
Tang, H.; and Ellis, K. 2022.
\newblock From Perception to Programs: Regularize, Overparameterize, and
  Amortize.
\newblock In \emph{Proceedings of the 6th ACM SIGPLAN International Symposium
  on Machine Programming}, 30–39.

\bibitem[{Valkov et~al.(2018)Valkov, Chaudhari, Srivastava, Sutton, and
  Chaudhuri}]{houdini}
Valkov, L.; Chaudhari, D.; Srivastava, A.; Sutton, C.; and Chaudhuri, S. 2018.
\newblock HOUDINI: Lifelong Learning as Program Synthesis.
\newblock In \emph{NeurIPS}.

\bibitem[{Wager and Athey(2018)}]{causalforest}
Wager, S.; and Athey, S. 2018.
\newblock Estimation and inference of heterogeneous treatment effects using
  random forests.
\newblock \emph{Journal of the American Statistical Association}, 113(523):
  1228--1242.

\bibitem[{Winskel(1993)}]{backusnaur}
Winskel, G. 1993.
\newblock \emph{The Formal Semantics of Programming Languages: An
  Introduction}.
\newblock Cambridge, MA, USA: MIT Press.

\bibitem[{Yao et~al.(2018)Yao, Li, Li, Huai, Gao, and Zhang}]{repr}
Yao, L.; Li, S.; Li, Y.; Huai, M.; Gao, J.; and Zhang, A. 2018.
\newblock Representation Learning for Treatment Effect Estimation from
  Observational Data.
\newblock In \emph{NeurIPS}.

\bibitem[{Yoon, Jordon, and van~der Schaar(2018)}]{yoon2018ganite}
Yoon, J.; Jordon, J.; and van~der Schaar, M. 2018.
\newblock {GANITE:} Estimation of Individualized Treatment Effects using
  Generative Adversarial Nets.
\newblock In \emph{ICLR}.

\bibitem[{Zhang, Liu, and Li(2021)}]{disent}
Zhang, W.; Liu, L.; and Li, J. 2021.
\newblock Treatment Effect Estimation with Disentangled Latent Factors.
\newblock In \emph{AAAI}.

\bibitem[{Zoph and Le(2017)}]{nasrl}
Zoph, B.; and Le, Q. 2017.
\newblock Neural Architecture Search with Reinforcement Learning.
\newblock In \emph{ICLR}.

\end{thebibliography}
